%% file: main.tex
\documentclass[runningheads]{llncs}
\usepackage[T1]{fontenc}
\usepackage{graphicx}
\usepackage{booktabs}
\usepackage[misc]{ifsym}
\newcommand{\corr}{(\Letter)}
\usepackage{mwe}

\usepackage[usenames,dvipsnames,table,xcdraw]{xcolor}
\usepackage{float}
\usepackage{multirow}
\usepackage{nicematrix}
\usepackage{comment}
\usepackage{algorithm}
\usepackage{algorithmic}
\usepackage{amsfonts}
\usepackage{amsmath,amssymb}

\newcommand{\innerproductcomma}[2]{\langle #1, #2 \rangle}

\usepackage{booktabs}
\usepackage{subcaption}
\usepackage{appendix}

\usepackage{hyperref}

\begin{document}

\title{Continuous Geometry-Aware Graph Diffusion via Hyperbolic Neural PDE}


\author{Jiaxu Liu\inst{1}\orcidID{0000-0001-8737-3106} \and
Xinping Yi\inst{2} \corr \and
Sihao Wu\inst{1} \and
Xiangyu Yin\inst{1} \and
Tianle Zhang \inst{1} \and
Xiaowei Huang \inst{1} \and
Shi Jin \inst{2}
}

\authorrunning{J. Liu et al.}

\institute{
Department of Computer Science, University of Liverpool, Liverpool, UK \and
National Mobile Communications Research Laboratory, Southeast University, Nanjing, China \\
\email{\{firstname.lastname\}@liverpool.ac.uk, \{xyi,jinshi\}@seu.edu.cn}
}

\tocauthor{Jiaxu~Liu,Xinping~Yi,Sihao~Wu,Xiangyu~Yin,Tianle~Zhang,Xiaowei~Huang,Shi~Jin}
\toctitle{Continuous Geometry-Aware Graph Diffusion via Hyperbolic Neural PDE}

\maketitle              

\begin{abstract}
While Hyperbolic Graph Neural Network (HGNN) has recently emerged as a powerful tool dealing with hierarchical graph data, the limitations of scalability and efficiency hinder itself from generalizing to deep models. In this paper, by envisioning depth as a continuous-time embedding evolution, we decouple the HGNN and reframe the information propagation as a partial differential equation, letting node-wise attention undertake the role of diffusivity within the Hyperbolic Neural PDE (HPDE). By introducing theoretical principles \textit{e.g.,} field and flow, gradient, divergence, and diffusivity on a non-Euclidean manifold for HPDE integration, we discuss both implicit and explicit discretization schemes to formulate numerical HPDE solvers. Further, we propose the Hyperbolic Graph Diffusion Equation (HGDE) -- a flexible vector flow function that can be integrated to obtain expressive hyperbolic node embeddings. By analyzing potential energy decay of embeddings, we demonstrate that HGDE is capable of modeling both low- and high-order proximity with the benefit of local-global diffusivity functions. Experiments on node classification and link prediction and image-text classification tasks verify the superiority of the proposed method, which consistently outperforms various competitive models by a significant margin.

\end{abstract}

\input{Content/introduction}
\input{Content/Preliminary}

\input{Content/HypODE}

\input{Content/HypGPDE}

\input{Content/Experiment}

\input{Content/Conclusion}

\begin{credits}
\subsubsection{\ackname} XH is financially supported by the U.K. EPSRC through End-to-End Conceptual Guarding of Neural Architectures [EP/T026995/1]. JL is supported by Liverpool-CSC scholarship [202208890034].

\end{credits}
%
%
%
\bibliographystyle{splncs04}
\bibliography{mybibliography}

\clearpage
\appendix
\input{Content/Appendices/AppA}
\clearpage
\input{Content/Appendices/AppB}

\clearpage
\input{Content/Appendices/AppC}
\clearpage
\input{Content/Appendices/AppD}

\end{document}

%% file: Content/introduction.tex
\section{Introduction}
Graphs play a vital role in various disciplines, including social network analysis \cite{clauset2008hierarchical}, bioinformatics \cite{zitnik2019evolution}, and computer vision \cite{yan2018spatial}. The advent of Graph Neural Networks (GNNs, \cite{kipf2016semi}) has significantly enhanced the analysis of these structures to capture complex relationships between nodes in a graph. However, traditional GNNs operate within the borders of Euclidean space, which may not be sufficiently expressive for data with inherent hierarchical or complex structures. To improve, this paper delves into the realm of hyperbolic geometry, a Riemannian manifold demonstrated to be particularly effective for embedding hierarchical data \cite{gromov1987hyperbolic,hamann2018tree}. We focus on the development of HGNNs \cite{chami2019hyperbolic}, which leverage the unique properties of hyperbolic space to enhance the embedding of GNNs.


The principal challenge confronted by HGNNs is their architectural design, which primarily consists of combinations of aggregation and transformation within layers. This fusion presents a unique problem, particularly the difficulty of training attention weights and manifold parameters (\textit{e.g.}, curvature of the hyperbolic manifold) layer-wise in a \textbf{deeply} layered scheme. With such challenge, we pose our initial questions: \textbf{Q1}: \textit{Considering hyperbolic space slows down layer-wise attention and propagation \cite{choudhary2022towards,liu2024deephgcn}, how to develop a deeply-layered attentive HGNN?} \textbf{Q2}: \textit{How to incorporate high-order info to benefit a deeply layered scheme?} \textbf{Q3}: \textit{Deep GNNs suffer from embedding smoothing, how should the node smoothness be measured when there is no defined metric for hyperbolic smoothness. And how to tackle over-smoothing within hyperbolic manifold constraints?}


Motivated by above questions, in this paper, we propose to decouple the functions within layers of HGNNs so as to deal with each of them separately. Unlike traditional decoupling-GNN approaches \cite{gasteiger2018predict,wu2019simplifying} that aggregate all information from the neighbors, we view information propagation as a distillation process, such that unimportant information is filtered out and significant information is weighted and contributes to the continuous variation of embeddings. More explicitly, by letting the transformation layer manifest as an encoder-decoder scheme, the aggregation layer is re-envisioned to solve the partial differential equation (Neural ODE/PDE, \cite{chen2018neural}) - essentially, the graph diffusion equation \cite{chamberlain2021grand} in hyperbolic space, which essentially simulates an infinitely deep HGNN with single layer parameters. In specific, in response to \textbf{Q1}, we consider the PDE reformulation and developed Hyperbolic-PDE (HPDE) solvers, which only leverage single-layer parameters. To answer \textbf{Q2}, we formulate the Hyperbolic Graph Diffusion Equation (HGDE), a low-high order vector flow function that can be integrated by HPDE. Tackling \textbf{Q3}, we firstly introduce the hyperbolic adaptation of Dirichelet energy and augmented HGDE with a hyperbolic residual, powered by Poincar\'e midpoint. Deconstructions above introduce extensive mathematical principles, including for instance: manifold vector field, flow, gradient, divergence, diffusivity, numerical HPDE solvers and hyperbolic residuals for bounding embedding energy decay. Through these concepts, we open new pathways to fully exploit the unique potential of hyperbolic space in the contextual analysis of graph-based data. In summary, the contributions of this paper are listed as follows.

\textbf{(I)} We present the geometric intuition for designing projective numerical integration methods that solve hyperbolic ODE/PDE, and examine the connection to Riemannian gradient descent methods. Focusing on fixed-grid solvers, we derive both hyperbolic generalizations of explicit schemes (Euler, Runge-Kutta) and implicit schemes (Adams-Moulton).

\textbf{(II)} We formulate the HGDE, which acts as the \textit{vector flow} of the HPDE, and thereby induces concepts such as gradient, divergence and diffusivity within HGDE. The proposed framework is flexible and efficient for generating expressive (endowed by the depth) hyperbolic graph embeddings.

\textbf{(III)} We instantiate the diffusivity function as a mixed-order multi-head attention to account for both homophilic (local) and heterophilic (global) relations. Besides, we introduce hyperbolic residual technique to benefit the optimization and prevent over-smoothing.

Through extensive experiments and comparison with the state-of-the-art on multiple real-world datasets, we show that HGDE framework can not only learn comparably high-quality node embeddings as Euclidean models on non-hierarchical datasets, but outperform all compared hyperbolic models variants on highly-hierarchical datasets with improved efficiency and accuracy. \textcolor{Purple}{The code and appendix can be found in \url{https://github.com/ljxw88/HyperbolicGDE}}.


%% file: Content/Preliminary.tex
\section{Preliminaries}
\subsubsection{Riemannian Geometry and Hyperbolic Space} A Riemannian manifold $\mathcal{M}$ of $n$-dimension is a topological space associated with a metric tensor $g$, denoted as $(\mathcal{M}, g)$, which extends curved surfaces to higher dimensions and can be locally approximated by $\mathbb{R}^n$. At any point $\mathbf{x}\in \mathcal{M}$, the tangent space $\mathcal{T}_\mathbf{x}\mathcal{M}\cong \mathbb{R}^n$ represents the first-order approximation of a small perturbation around $\mathbf{x}$, isomorphic to Euclidean space. The Riemannian metric $g$ on the manifold determines a smoothly varying positive definite inner product on the tangent space, enabling the definition of diverse properties \textit{e.g.} geodesic length, angles, and curvature.

The hyperbolic space $\mathbb{H}^n$ is a smooth Riemannian manifold with a constant negative sectional curvature $\kappa < 0$. Its coordinates can be represented via various isometric models. \cite{beltrami1868teoria} established the equivalence of hyperbolic and Euclidean geometry through the utilization of the $n$-dimensional \textit{Poincar\'e ball model}, which equips an open ball $\mathbb{D}^n_\kappa = (\mathcal{D}^n_\kappa, g^{\mathbb{D}})$, with point set $\mathcal{D}^n_\kappa = \{\mathbf{x} \in \mathbb{R}^n : \|\mathbf{x}\| < -\frac{1}{\kappa}\}$ and Riemannian metric $g^{\mathbb{D}}_\mathbf{x} = (\lambda_\mathbf{x}^\kappa)^2 \mathbf{I}_n$, where the conformal factor $\lambda_\mathbf{x}^\kappa = \frac{2}{1+\kappa\|\mathbf{x}\|^2}$. The Poincar\'e metric tensor induces various geometric properties \textit{e.g.} distances $d^\kappa_\mathbb{D}(\mathbf{x}, \mathbf{y})$, inner products $\innerproductcomma{\mathbf{u}}{\mathbf{v}}_\mathbf{x}^\kappa$, geodesics $\gamma_{\mathbf{x}\to \mathbf{y}}(t)$ and more \cite{nickel2017poincare}. Geodesics also induce the definition of \textit{exponential} and \textit{logarithmic} maps \cite{ganea2018entailment}. At point $\mathbf{x}\in\mathbb{D}^n_\kappa$, the exponential map $\exp_\mathbf{x}^\kappa: \mathcal{T}_\mathbf{x}\mathbb{D}_\kappa^n \to \mathbb{D}_\kappa^n$ essentially maps a small perturbation of $\mathbf{x}$ by $\mathbf{v}\in \mathcal{T}_\mathbf{x}\mathbb{D}_\kappa^n$ to $\exp_\mathbf{x}^\kappa (\mathbf{v})\in \mathbb{D}_\kappa^n$, so that $t\in[0,1]: \exp_\mathbf{x}^\kappa (t\mathbf{v})$ is the geodesic from $\mathbf{x}$ to $\exp_\mathbf{x}^\kappa (\mathbf{v})$. The logarithmic map $\log_\mathbf{x}^\kappa: \mathbb{D}_\kappa^n \to \mathcal{T}_\mathbf{x}\mathbb{D}_\kappa^n$ is defined as the inverse of $\exp_\mathbf{x}^\kappa$. Finally, the parallel transport $\mathcal{PT}_{\mathbf{x}\to\mathbf{y}}: \mathcal{T}_\mathbf{x}\mathbb{D}_\kappa^n \to \mathcal{T}_\mathbf{y}\mathbb{D}_\kappa^n$ moves a tangent vector $\mathbf{v}\in \mathcal{T}_\mathbf{x}\mathbb{D}_\kappa^n$ along the geodesic to $\mathcal{T}_\mathbf{y}\mathbb{D}_\kappa^n$ while preserving the metric tensor. For closed-form expression of above operations, please refer to Appendix~B.

\subsubsection{Diffusion Equations} The process of generating representations of individual data points through information flows can be characterized by an an-isotropic \textit{diffusion} process, a concept borrowed from physics used to describe heat diffusion on Riemannian manifold. Denote the manifold as $\mathcal{M}$, and let $z(t)$ denote a family of functions on $\mathcal{M}\times [0, \infty)$ and $z(u, t)$ be the density at location $u\in\mathcal{M}$ and times $t$. The general framework of diffusion equations is expressed as a PDE
\begin{align}
    \partial z (u,t)/\partial t = \mathrm{div}( a(z(u, t)) \nabla z(u, t) ), \quad t>0 \label{eq:general-framework-pde}
\end{align}
where $a(\cdot)$ defines the \textit{diffusivity} function controlling the diffusion strength between any location pair at time $t$. The gradient operator $\nabla: \mathcal{M}\to \mathcal{T}\mathcal{M}$ describes the steepest change at point $u\in \mathcal{M}$. $\mathrm{div}(\cdot):\mathcal{T}\mathcal{M}\to \mathcal{M}$ is the divergence operator that summarizes the flow of the diffusivity-scaled vector field $(a(\cdot)\nabla)$. Eq.~(\ref{eq:general-framework-pde}) can be physically viewed as a variation of heat based on time at the location $i$, identical to the heat that flows through that point from the surrounding areas. 

\subsubsection{Graph Diffusion Equation} Let $\mathcal{G}=(\mathcal{V}, \mathcal{E})$ denote an undirected graph with the node set $\mathcal{V}$ and the edge set $\mathcal{E}$. Let $\mathbf{x} = \{\mathbf{x}_i\in \mathbb{R}^d\}_{i=1}^{|\mathcal{V}|}$ be the node features and $\mathbf{z}(t)$ be node embeddings at time $t$. Process Eq.~(\ref{eq:general-framework-pde}) can be re-written as
\begin{align}
    \partial \mathbf{z}_i (t)/\partial t = \mathrm{div}( \mathbf{A}(\mathbf{z}(t)) \nabla \mathbf{z}_i(t) ), \label{eq:diff-process-general-graph}
\end{align}
where $\mathbf{A}$ is generally realised by a time-independent $n\times n$ attention matrix \cite{chamberlain2021grand,chamberlain2021beltrami}, consistent with the flow of \textit{heat flux} in/out node $i$. The formulation of Eq.~(\ref{eq:diff-process-general-graph}) as a PDE allows leveraging vast existing numerical integration methods to solve the continuous dynamics.

%% file: Content/HypODE.tex
\begin{figure}[t]
\centering
\includegraphics[width=1.0\textwidth]{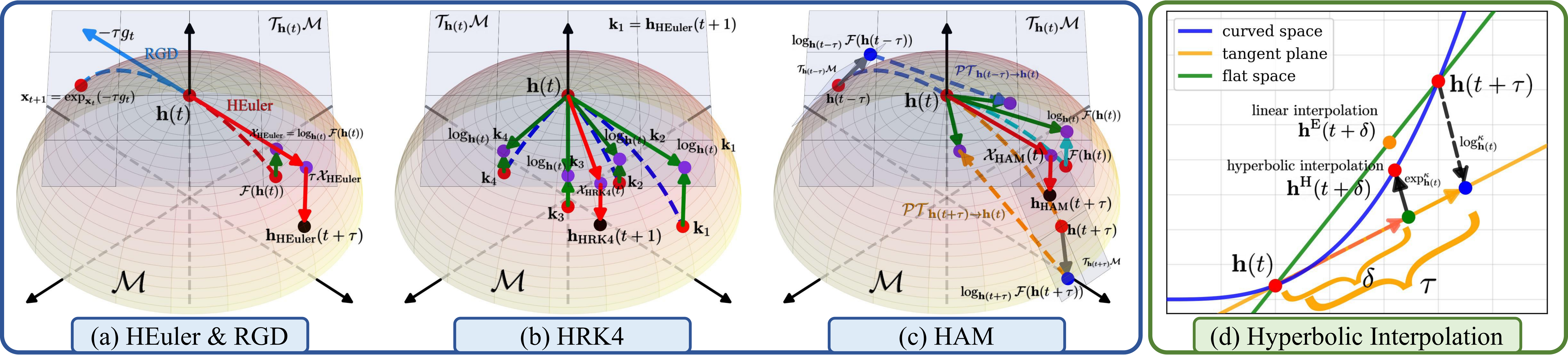}
\caption{\textit{(a-c)} Illustration of various numerical integration methods with comparison to RGD. In each time-step, an explicit scheme calibrates the vector field within only the tangent space of time $t$, while an implicit scheme requires multiple tangent spaces to estimate future slopes, thus requiring parallel transport for aligning the directions of vectors in different spaces. \textit{(d)} Illustration of hyperbolic interpolation method.}
\label{Fig.hypint}
\end{figure}

\section{Hyperbolic Numerical Integrators}
\label{sec:hyperbolic-solvers}
Consider the continuous form of ODE/PDE specified by a neural network parameterized by $\theta$, expressed as
\begin{equation}
    d \mathbf{h}(t)/d t = f_{\theta}(\mathbf{h}(t), t), \quad \mathbf{h}(0) = \mathbf{h}_0\label{eq:ode-nn}
\end{equation}
where the time step $t=[0, T]$. Eq.~(\ref{eq:ode-nn}) essentially tells that the \textit{rate of change} of $\mathbf{h}(t)\in \mathbb{R}^n$ at each time step is given by the vector field $f_\theta: \mathbb{R}^n\times\mathbb{R} \to \mathbb{R}^n$. Eq.~(\ref{eq:ode-nn}) is integrated to obtain $\mathbf{h}(T)$. In our context, we are interested in formulating a PDE recipe that is aware of hyperbolic geometry.
\begin{definition}
\label{def:manifold-vector-field}
    A time-dependent \textbf{manifold vector field} is a mapping $\mathcal{X}:\mathcal{M}\times \mathbb{R}\to\mathcal{TM}$, which assigns each point in $\mathcal{M}$ at $t$ a tangent vector. The particle’s time-evolution according to $\mathcal{X}$ is then given by the following PDE
    \begin{align}
        d \mathbf{h}(t)/d t = \mathcal{X}_\theta(\mathbf{h}(t), t).\label{eq:hyperbolic-ode-continous-form}
    \end{align}
\end{definition}
\begin{definition}
\label{def:manifold-flow}
    A \textbf{vector flow} is a mapping generated by vector field, \textit{i.e.} $\mathcal{F}\equiv\pi_{}(\mathcal{X})$, where $\pi:\mathcal{M}\to\mathcal{M}$ is a smooth projection of vector field to manifold of their local coordinates. Vice versa, if $\pi$ is a diffeomorphism, then $\mathcal{X}\equiv \pi^{-1}(\mathcal{F})$.
\end{definition}
In hyperbolic geometry, where $\pi$ and $\pi^{-1}$ are properly defined $\exp$ and $\log$ maps, our concern lies in the particle's location on the manifold subsequent to integration, \textit{i.e.} integrate through the path defined by flow $\mathcal{F}$. This can be achieved via the spirit of projective method \cite{hairer8solving}. In the following, we derive numerical solvers for estimating the integral of field $\mathcal{X}$ or flow $\mathcal{F}$ \textit{w.r.t.} time $t$ using, respectively, the explicit and implicit schemes.

\subsection{Hyperbolic Projective Explicit Scheme}
In an explicit scheme, the state at the next time step is computed directly from the current state and its derivatives. In this part, we derive the hyperbolic generalization of the explicit scheme. To illustrate high-level ideas, we introduce both \textit{single step method} and \textit{multi-step method}. We also discuss the geometric intuition and strong analogy between one-step explicit scheme and Riemannian gradient descent (RGD).

\subsubsection{H-Explicit Euler (HEuler)}
Consider a small time step $\tau$. Iteratively, we seek an approximation for $\mathbf{h}(t + \tau)$ based on $\mathbf{h}(t)$ and vector field $f(\cdot)$. In Euclidean space, the explicit Euler method is written as
\begin{equation}
    \mathbf{h}(t+\tau) \approx \mathbf{h}(t) + \tau f_{\theta}(\mathbf{h}(t), t),\label{Eulidean-euler}
\end{equation}
which is a discrete version of Eq.~(\ref{eq:ode-nn}). Similarly in hyperbolic space, we discretize Eq.~(\ref{eq:hyperbolic-ode-continous-form}), and have the stepping function formulated as
\begin{align}
     \mathbf{h}_{\mathrm{HEuler}}(t+\tau) = \exp^\kappa_{\mathbf{h}(t)} (\tau \mathcal{X}_{\mathrm{HEuler}}(t)),\label{hyperbolic-euler-1}
\end{align}
where the vector field $\mathcal{X}$ gives the direction at time $t$ according to flow $\mathcal{F}_\theta^\kappa$
\begin{equation}
    \mathcal{X}_{\mathrm{HEuler}}(t) = \log^\kappa_{\mathbf{h}(t)}(\mathcal{F}_{\theta}^\kappa(\mathbf{h}(t), t)) \in \mathcal{T}_{\mathbf{h}(t)}\mathbb{D}_\kappa^n. \label{hyperbolic-euler-2}
\end{equation}

\textit{\color{Blue}{Geometric Intuition}} The equation in Eq.(\ref{Eulidean-euler}) signifies a transition from $\mathbf{h}(t)$ in the direction of $f$ by a distance proportional to $\tau$. In hyperbolic space, where $\mathbf{h}(t)\in\mathbb{D}^n_{\kappa}$ and we presume $\mathcal{X}^{\kappa}: \mathbb{D}^n_{\kappa} \to \mathcal{T}\mathbb{D}^n_{\kappa}$, the transition follows the geodesic dictated by the direction of $\mathcal{X}^\kappa$. Recall the definition of exponential map: given $x\in \mathbb{D}_\kappa$, $\exp^\kappa_x(v)$ takes $v\in\mathcal{T}_x\mathbb{D}_\kappa$ and returns a point in $\mathbb{D}_\kappa$ reached by moving from $x$ along the geodesic determined by the tangent vector $v$. Thus Eq.(\ref{hyperbolic-euler-1}-\ref{hyperbolic-euler-2}) can be essentially viewed as a geometric transportation of points on manifold along the curve defined by $\mathcal{F}$.

\textit{\color{Brown}{Connection to RGD}} As visualized in Fig.~\ref{Fig.hypint}(a), the explicit Euler can be viewed as reversed RGD, where the direction $\mathcal{X}_\mathrm{HEuler}(t)$ plays similar role as the Riemannian gradient $g_t$ at $\mathbf{h}(t)$. Similar to RGD, when $(\mathcal{M}, \rho)$ is Euclidean space $(\mathbb{R}^n, \mathbf{I}_n)$, then Eq.~(\ref{hyperbolic-euler-1}) converges to Eq.~(\ref{Eulidean-euler}) since we have $\exp_h^\kappa(v) \xrightarrow[]{\kappa\to0} h+v$. This property is useful on developing higher-order integrators.

\subsubsection{H-Runge-Kutta (HRK)}
With a similar geometric intuition, we derive the hyperbolic extension of the Runge-Kutta method. Define the $s$-order HRK stepping function
\begin{equation}
    \mathbf{h}_{\mathrm{HRK}}(t + \tau) = \exp^\kappa_{\mathbf{h}(t)} (\tau \mathcal{X}_{\mathrm{HRK}}(t)),\label{hyperbolic-rk4-1}
\end{equation}
where the vector field is estimated by
\begin{align}
    &\mathcal{X}_{\mathrm{HRK}}(t) = \left(\sum_{i=1}^s \phi_i \log_{\mathbf{h}(t)}^\kappa (\mathbf{k}_i)\right) / {\sum_{i=1}^s \phi_i}.\label{eq:vector-flow-of-rk}
\end{align}
In Eq.~(\ref{eq:vector-flow-of-rk}), $\mathbf{k}$ denotes the vector flow functions, $\{\phi_i\}$ are coefficients determined by the order. Specifically for $4$th order Runge-Kutta (HRK4), we have $\{\phi_{1\dots 4}\}=\{1, 3, 3, 1\}$ derived from Taylor series expansion as in \cite{chen2018neural}. The vector flows $\mathbf{k}_{1\dots 4}$ are respectively formulated by
\begin{align}
    & \mathbf{k}_1 = \mathbf{h}_{\mathrm{HEuler}}(t+\tau),\qquad\text{(Eq.~(\ref{hyperbolic-euler-1}))}\\
    & \mathbf{k}_2 = \mathcal{F}_{\theta}^\kappa(\exp^\kappa_{\mathbf{h}(t)} (\tau \mathcal{X}_{\mathbf{k}_2}), t + \tau/3), \nonumber
    \text{ where } \mathcal{X}_{\mathbf{k}_2} = \log^\kappa_{\mathbf{h}(t)}(\mathbf{k}_1)/3. \nonumber\\
    & \mathbf{k}_3 = \mathcal{F}_{\theta}^{\kappa}(\exp^\kappa_{\mathbf{h}(t)}(\tau \mathcal{X}_{\mathbf{k}_3}), t + 2\tau/3), 
     \text{ where } \mathcal{X}_{\mathbf{k}_3} = \log^\kappa_{\mathbf{h}(t)}(\mathbf{k}_2) - \log^\kappa_{\mathbf{h}(t)}(\mathbf{k}_1)/3.  \nonumber\\
    & \mathbf{k}_4 = \mathcal{F}_{\theta}^\kappa (\exp^\kappa_{\mathbf{h}(t)}(\tau \mathcal{X}_{\mathbf{k}_4}), t+\tau),
    \text{ where } \mathcal{X}_{\mathbf{k}_4} = \log^\kappa_{\mathbf{h}(t)}(\mathbf{k}_1) -  \log^\kappa_{\mathbf{h}(t)}(\mathbf{k}_2) + \log^\kappa_{\mathbf{h}(t)}(\mathbf{k}_3). \nonumber
\end{align}
As illustrated in Fig.~\ref{Fig.hypint}(b), this method approximates the solution to the PDE within a small interval, considering not only the derivative at the initial time (as in Eq.~(\ref{Eulidean-euler})), but also at intermediate points and the end of the interval.

\subsection{Hyperbolic Projective Implicit Scheme}
In an implicit scheme, the state of the next iteration is computed by incorporating its own value. This requires solving a linear system to obtain $\mathbf{h}(t+\tau)$ based on $\mathbf{h}(t)$. In below, we illustrate a hyperbolic generalization of the implicit solver. 

\subsubsection{H-Implicit Adams–Moulton (HAM)}
Adams numerical integration methods are introduced as families of multi-step methods. With order $s=0$, Adams methods are identical to the Euler's method. Principally, there are two types of Adams methods, namely, Adams–Bashforth (explicit) and Adams–Moulton (implicit). Our emphasis is on the latter.

The implicit nature of AM requires the initialization of first several steps with a different method. We use the hyperbolic Runge-Kutta (Eq.~(\ref{hyperbolic-rk4-1})) for initialization. With the input $\mathbf{h}(t)\in \mathbb{D}_\kappa^n$ and flow $\mathcal{F}^\kappa$, define the warm up
\begin{align}
    &\mathbf{h}_{\mathrm{HAM}}(i\tau) = \mathbf{h}_\mathrm{HRK4}(i\tau), \quad 0\le i < s_\mathrm{min}\label{eq:init-ham}
\end{align}
where $s_\mathrm{min}$ is the min order. During the whole warm up process, we maintain a queue $\mathbf{q}$ of tangent vectors and the points spanning the tangent space. In each time step of Eq.~(\ref{eq:init-ham}), we push $[q_0 = \mathcal{X}_{\mathrm{RK4}}(i\tau) , q_1=\mathbf{h}(i\tau) ]$ to the head of $\mathbf{q}$. When $\mathrm{len}(\mathbf{q}) \ge s_\mathrm{min}$, we start the time-stepping
\begin{align}
    \mathbf{h}_{\mathrm{HAM}}(t+\tau) = \exp^\kappa_{\mathbf{h}(t)} (\tau \mathcal{X}_{\mathrm{HAM}}(t)), \label{eq:hyperbolic-implicit-adam-1}
\end{align}
where the vector field is expressed as
\begin{align}
    \mathcal{X}_{\mathrm{HAM}}(t) = &\phi_0\mathcal{PT}_{\mathbf{h}(t+\tau) \to \mathbf{h}(t)} ( \log^\kappa_{\mathbf{h}(t+\tau)} ( \mathcal{F}^\kappa_\theta ( \mathbf{h}(t+\tau), t+\tau ) ) )\nonumber \\
    &+ \sum_{i=1}^s \phi_i \mathcal{PT}_{\mathbf{q}_{i, 1} \to \mathbf{h}(t)} ( \mathbf{q}_{i, 0} ).
\end{align}
The order $s=\min(\mathrm{len}(\mathbf{q}), s_\mathrm{max})$, $\{\phi_i\}$ are coefficients determined by the order, which are typically within a pre-defined look-up table. As illustrated in Fig.~\ref{Fig.hypint}(c), since the reference point $\mathbf{h}(t)$'s stored in $\mathbf{q}$ are different, the parallel transport $\mathcal{PT}$ is leveraged for aligning tangent spaces for different slopes. When $\mathbf{h}_{\mathrm{HAM}}(t+\tau)$ is accepted as converged, $[\log^\kappa_{\mathbf{h}(t)}(\mathbf{h}_{\mathrm{HAM}}(t+\tau) ), \mathbf{h}(t) ]$ is pushed to $\mathbf{q}$ for the next iteration and the last element is popped if $\mathrm{len}(\mathbf{q})$ reaches $s_\mathrm{max}$. We refer readers to Appendix~{C} for detailed explanation of the algorithms.

\subsection{Interpolation on Curved Space}
Fixed grid PDE solvers typically use their own internal step sizes $\tau$ to advance the solution of the PDE. For certain time step $t$, given $\mathbf{h}(t)$ and $\mathbf{h}(t+\tau)$, we may want to obtain the solution at time point $t+\delta$ where $0<\delta<\tau$. Since $\delta$ does not lie on the grid defined by $\{0, \tau\}$, interpolation methods are invoked to estimate $\mathbf{h}(t+\delta)$. For hyperbolic geometry that $\mathbf{h}\in \mathbb{D}^n_\kappa$, define the interpolation
\begin{align}
    \mathbf{h}(t+\delta) = \exp_{\mathbf{h}(t)}^\kappa \left({\delta} \log^\kappa_{\mathbf{h}(t)}(\mathbf{h}(t+\tau)) / {\tau}\right).  \label{eq:hyperbolic-linear-interpolation}
\end{align}
\begin{proposition}[proved in Appendix~D]
    For any step size $0<\delta<\tau$, the interpolation $\mathbf{h}(t+\delta)$ via Eq.~(\ref{eq:hyperbolic-linear-interpolation}) is on the geodesic between $\mathbf{h}(t)$ and $\mathbf{h}(t+\tau)$ on the manifold, and $\frac{d_\mathbb{D}^\kappa(\mathbf{h}(t), \mathbf{h}(t+\delta))}{ d_\mathbb{D}^\kappa(\mathbf{h}(t), \mathbf{h}(t+\tau)) }=\frac{\delta}{\tau}$ where $d_\mathbb{D}^\kappa$ is the geodesic length. 
\label{prop:proportional-geodesic-inter}
\end{proposition}

%% file: Content/HypGPDE.tex
\section{Diffusing Graphs in Hyperbolic Space}

\subsection{Hyperbolic Graph Diffusion Equation}
We study the diffusion process of graphs with node representation residing in hyperbolic geometry. Given the diffusion time $t\in [0,T]$, embedding space $\mathbb{D}^d_{\kappa_t}$ with learnable curvature $\kappa_t$ at time $t$, node embedding $\mathbf{z}_{*}(t) \in \mathbb{D}^d_{\kappa_t}$ and $\mathcal{C(\cdot)}$ being the correlated coordinates of certain node, we formulate the vector flow $\mathcal{F}_\theta^\kappa$ of the $i$th representation as
\begin{align}
    \underbrace{\exp_{\mathbf{z}_i(t)}^{\kappa_t} \bigg( \sigma \bigg[ \sum_{j\in\mathcal{C}(i)} }_\text{divergence} \underbrace{a(\mathbf{z}_i(t), \mathbf{z}_j(t))}_\text{diffusivity} \underbrace{ \log_{\mathbf{z}_i(t)}^{\kappa_t} (\mathbf{z}_j(t)) }_\text{gradient}  \bigg] \bigg), \label{eq:hyperbolic-graph-diffusion-equation}
\end{align}
where $\sigma$ can be either identity/non-linear activation. With initial state encoded by learnable feature transformation $\psi$, \textit{i.e.} $\mathbf{z}(0) = \psi(\mathbf{x})\in \mathbb{D}^d_\kappa$, the final state can be numerically estimated by our proposed HPDE integrators, \textit{i.e.} $\mathbf{z}_i(T) = \mathrm{HPDESolve}(\mathbf{z}_i(0), \frac{\partial \mathbf{z}_i(t)}{\partial t}, 0, T)$. In matrix form, the vector flow is expressed as
\begin{align}
    \mathcal{F}^\kappa_\theta(\mathbf{z}(t), t) =  \exp^{\kappa_t}_{\mathbf{z}(t)} \left( \sigma \big[ \mathbf{S}(\mathbf{z}(t))\nabla\mathbf{z}(t) \big]\right),\label{eq:hyperbolic-graph-diffusion-equation-matrix}
\end{align}
where $\mathbf{S}(\mathbf{z}(t)) = (a(\mathbf{z}_i(t),\mathbf{z}_j(t)))$ is a normalized $|\mathcal{V}|\times |\mathcal{V}|$ similarity matrix, and $\nabla\mathbf{z}(t))_{i,j} := \log_{\mathbf{z}_i(t)}^{\kappa_t} (\mathbf{z}_j(t)$. In below, we discuss the key ingredients of Eq.~(\ref{eq:hyperbolic-graph-diffusion-equation}-\ref{eq:hyperbolic-graph-diffusion-equation-matrix}).


\begin{figure}[t]
\centering
\includegraphics[width=1.0\textwidth]{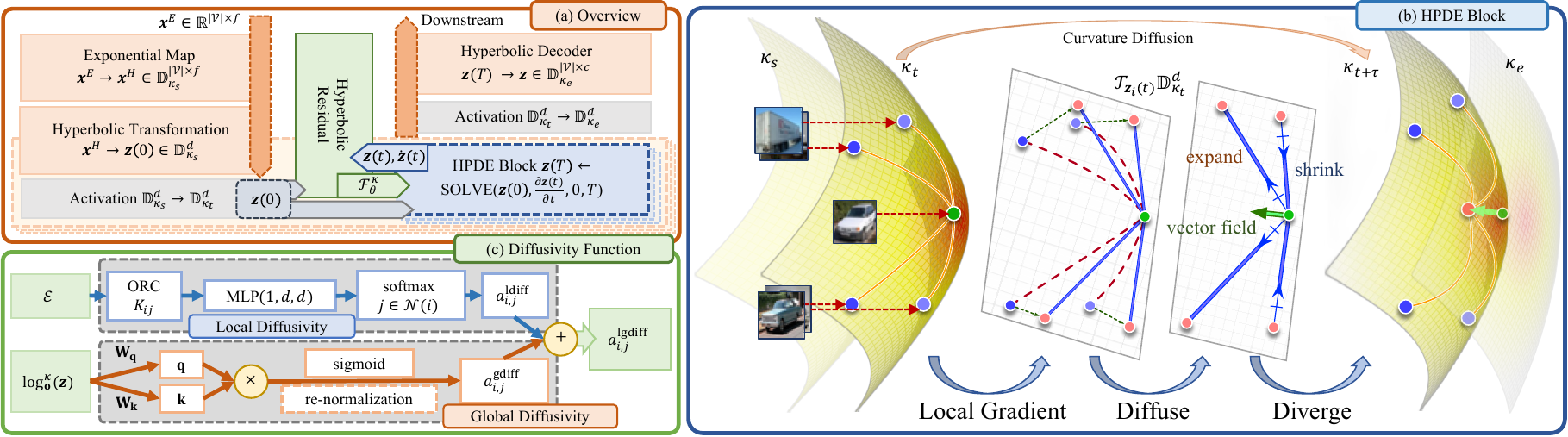}
\caption{Schematic of HGDE. \textit{(a)} The pipeline of our method includes hyperbolic projection, feature transformation, and HPDE block that integrates the GDE. After that, a decoder is applied to the embeddings for specific downstream tasks. \textit{(b)} The visualization of the diffusion process within the HPDE block: first, map local gradients of $\mathbf{z}_i$ onto the tangent space, calculate the diffusivity, and diverge to obtain the vector flow (green arrow), then perform one-step integration on the manifold with the guidance of continuous curvature diffusion. \textit{(c)} The details of attention-powered local-global diffusivity function.}
\label{Fig.pipeline}
\end{figure}

\subsubsection{Gradient}
The gradient of a function $z(u,t)$ at location $u$ in a discrete space can be approximated as the difference between the function values at neighboring points. In graph space, let $\mathbf{z}_i$ and $\{\mathbf{z}_j\}_{j\in \mathcal{C}(i)}$, respectively, denote the target node and the correlated positions of $i$ that can be modeled by edge connectivity or self-attention. The graph diffusion process \cite{chamberlain2021grand,chamberlain2021beltrami} treats nodes as Euclidean representations, such that the analogy of gradient operator $(\nabla \mathbf{z}(t))_{i,j} : \mathbb{R}^d \to \mathbb{R}^d$ is expressed as $\mathbf{z}_j(t) - \mathbf{z}_i(t)$. However, when nodes are embedded in Riemannian manifolds, the gradient of a node is no longer the difference between itself and neighboring points. Instead, we take vectors in the tangent space at $\mathbf{z}_i$ that are obtained by taking the derivative of $\mathbf{z}$ in all possible directions, \textit{i.e.} $(\nabla \mathbf{z}(t))_{i,j}: \mathbb{D}_\kappa^d \to \mathcal{T}\mathbb{D}_\kappa^d$ that can be formulated as $\log_{\mathbf{z}_i(t)}^\kappa (\mathbf{z}_j(t))$. One recovers the discrete Euclidean gradient as the curvature $\kappa\to 0$.

\subsubsection{Diffusivity}
The diffusivity scales the gradient, with either isotropic or anisotropic behavior. For graph diffusion, the \textbf{isotropic} formula is presented by the normalized adjacency matrix \cite{kipf2016semi}, where $a_{i,j} = \frac{1}{\sqrt{d_i d_j}}$ \textit{iff.} $(i,j)\in\mathcal{E}$ and $d$ is the degree.

Alternatively, the \textbf{anisotropic} approach incorporates the attention mechanism \cite{vaswani2017attention} to account for the asymmetric relationship between pairs of nodes. This paper considers \textit{local}, \textit{global} and \textit{local-global} schemes based on structure information. Define the schemes
$$
\begin{cases}
   a^{\mathrm{ldiff}}_{i,j} = \mathrm{normalize}_{j\in \mathcal{N}(i)} \left(
    f_\theta \big(\mathbf{z}_i(t) , \mathbf{z}_j(t)\big)
    \right)  & \text{(local scheme)} \\
  a_{i,j}^\mathrm{gdiff} = \beta \mathrm{normalize}_{j\in \mathcal{V}}\left(
    g_\phi \big(\mathbf{z}_i(t) , \mathbf{z}_j(t)\big)
    \right) + \frac{1-\beta}{\sqrt{d_i d_j}} & \text{(global scheme)} \\
    a_{i,j}^\mathrm{lgdiff} = \beta \mathrm{normalize}_{j\in \mathcal{V}}\left(
    g_\phi \big(\mathbf{z}_i(t) , \mathbf{z}_j(t)\big)
    \right) + (1-\beta) a^{\mathrm{ldiff}}_{i,j} & \text{(local-global scheme)}
\end{cases}
$$
where $f_\theta/g_\phi$ are learnable functions that compute the diffusivity weight between node pair $(i, j)\in\mathcal{E}$. $\beta$ can be constant or trainable parameters that adjust the emphasis on homophilic (local attention) and heterophilic (high-order, global attention) relations. In comparison, the local attention scheme implicitly incorporates the graph information since only neighboring elements are considered based on $\mathcal{N}(i)$. Whereas for global attention, it neglects the graph topology and hence requires manual incorporation.

\textit{\color{OliveGreen}Low-order Local Diffusivity}. A straightforward approach is to leverage the formula of graph attention \cite{velivckovic2017graph}, which is extended to the hyperbolic space by \cite{chami2019hyperbolic}, where the weights are calculated tacitly in the tangent space. An alternative method to consider is the Oliver-Ricci Curvature (ORC) \cite{ollivier2009ricci} attention, introduced in \cite{ye2019curvature,yang2023kappahgcn} to drive message propagation. This approach is not limited by the non-Euclidean property of node feature, as it computes attention weight via the ORC value derived from the graph topology, thus allowing adoption without leveraging tangent space.

\textit{\color{OliveGreen}High-order Global Diffusivity}. Propagation of high-order node pairs results in exponentially increasing complexity compared to $f_\theta$. \cite{yun2019graph,wu2023difformer} introduced a series of scalable and efficient node-level transformers. With a similar notion in the hyperbolic space, we first project the embeddings onto the tangent space of the origin. Subsequently, the weights can be obtained using existing graph transformer architectures. We adopt energy-constrained transformers \cite{wu2023difformer} with a sigmoid kernel, which performs well in most scenarios. 

\textcolor{Red}{*} Fig.~\ref{Fig.pipeline}(c) presents the high-level schematic of \textit{diffuse}. The implementation and algorithmic details are delegated to Appendix~{C}.


\subsubsection{Divergence}
For simplicity, we assume any $\mathbf{x}_i\in \mathbb{R}^d$ to be scalar-valued. The divergence at a point $\mathbf{z}_i$ is a measure of how much the vector field $\mathcal{X}=\{\nabla \mathbf{z}(t))_{i,j}\}_{j\in \mathcal{C}(i)}$ is expanding or contracting at $\mathbf{z}_i$. In a Euclidean space, the divergence would indeed be the sum of the components of the gradient, \textit{i.e.}, \(\mathrm{div}_i = \sum_j (\nabla \mathbf{z}(t))_{i,j}\), producing a scalar (with dimensionality $\mathcal{T}_{\mathbf{z}_i}\mathbb{D}^d \cong\mathbb{R}^d$). In our context, we are interested in how $\mathbf{z}_i$ is varied in the manifold rather than in the tangent space; thus an exponential map is applied to the sum of gradients on $\mathcal{T}_{\mathbf{z}_i}\mathbb{D}^d$, giving \(\mathrm{div}_i = \exp_{\mathbf{z}_i(t)}(\sum_j a_{i,j}(\nabla \mathbf{z}(t))_{i,j})\). This also satisfies the form of $\mathcal{F}^\kappa_\theta$ in Def.~\ref{def:manifold-flow}, and thus can be numerically integrated through $\mathrm{HPDESolve}$.

\subsubsection{Continuous Curvature Diffusion}
Eq.~(\ref{eq:hyperbolic-graph-diffusion-equation-matrix}) implicitly guides the manifold towards its optimal geometry for embedding $\mathbf{z}(t)$ as the manifold parameter $\kappa_t$ also accumulates and is updated during backpropagation. Similar to the attention parameters $\theta$, we let $\kappa$ be time independent based on the assumption that $\lim_{\tau\to0}\frac{\kappa_{t+\tau} - \kappa_t}{\tau} = 0$.

\subsection{Convergence of Dirichlet Energy}
    
\begin{definition}
Given the node embedding $\{\mathbf{z}^{}_i  \in \mathbb{D}^{d}_\kappa\}_{i=1}^{|\mathcal{V}|}$, the hyperbolic Dirichlet energy is defined as
\begin{align}
\resizebox{0.8\hsize}{!}{%
$f^{\kappa}_\mathrm{DE}(\mathbf{z}) = \frac{1}{2} \sum_{(i,j)\in \mathcal{E}} {d_{\mathbb{D}}^\kappa} \left( 
\exp^\kappa_\mathbf{o}\left(\frac{\log_\mathbf{o}^\kappa(\mathbf{z}_i)}{\sqrt{1+d_i}}\right)  ,\exp^\kappa_\mathbf{o}\left(\frac{\log_\mathbf{o}^\kappa(\mathbf{z}_j)}{\sqrt{1+d_j}}\right)
\right)^2$
}
\label{eq:dirichlet-energy-definition-hyperbolic},
\end{align}
where $d_{i/j}$ denotes the node degree of node $i/j$. The distance $d_\mathbb{D}^\kappa(\mathbf{x}, \mathbf{y})$ between two points $\mathbf{x}, \mathbf{y}\in \mathbb{D}$ is the geodesic length; we detail the closed form expression in Appendix~{B}.
\label{def:dirichlet-energy-definition}
\end{definition}
Def.~\ref{def:dirichlet-energy-definition} introduces a node-similarity measure to quantify over-smoothness in hyperbolic space. $f^{\kappa}_\mathrm{DE}$ of node representation can be viewed as the weighted sum of distance between normalized node pairs. \cite[Prop.~4]{liu2024deephgcn} proved that hyperbolic energy $f^{\kappa}_\mathrm{DE}$ diminishes after message passing, and multiple aggregations result in converging towards zero energy, indicating reduced embedding expressiveness that could potentially cause over-smoothing. Also as proved in  \cite[Prop.~2]{zhang2022improving} that over-smoothing is an intrinsic property of first-order continuous GNN. In a continuous diffusion process, where each iteration can be viewed as a layer in HGNNs, as supported by Fig.~\ref{Fig.energy-convergence-ablation}, we also observe a convergence of hyperbolic Dirichlet energy of $\mathbf{z}(t)$ \textit{w.r.t.} time $t$.

\subsubsection{Residual-Empowered Flow}
Empirically, studies in multi-layer GNNs \cite{gasteiger2018predict,li2019deepgcns} demonstrated the efficacy of adding residual connections to the initial layer. It is also claimed in \cite{zhou2021dirichlet} that using residual connections for both initial and previous layers can prevent the Dirichlet energy from reaching a lower energy limit, thus avoiding over-smoothing. Building upon these studies, we define the hyperbolic residual empowered vector flow
\begin{equation}
    \mathcal{F}^\kappa_\theta(\mathbf{z}(t), t) = \mu^\kappa_\mathbb{D}\left(\{\dot{\mathbf{z}}(t), \mathbf{z}(t), \mathbf{z}(0)\}; \{\eta\}_{j=1}^J \right), \label{eq:residual-diffusion-hyperbolic}
\end{equation}
where $\dot{\mathbf{z}}(t) =  \exp^{\kappa_t}_{\mathbf{z}(t)} \left( \sigma \big[ \mathbf{S}(\mathbf{z}(t))\nabla\mathbf{z}(t) \big]\right)$ is the manifold dynamic as in Eq.~(\ref{eq:hyperbolic-graph-diffusion-equation-matrix}). $\{\eta\}_{j=1}^J$ are the weight coefficients. $\mu_\mathbb{D}^\kappa$ is the node-wise hyperbolic averaging. We instantiate it via \textit{M\"obius Gyromidpoint} \cite{ungar2008gyrovector} for its trade-off between computational cost and precision. Define
\begin{align}
    \mu_\mathbb{D}^\kappa (\{\mathbf{z}\}_{j=1}^J; \{\eta\}_{j=1}^J) = \left(\frac{1}{2}\otimes_\kappa \left( \frac{
    \sum_{j}\eta_j\lambda_{\mathbf{z}^{(j)}_i}^\kappa\mathbf{z}^{(j)}_i
    }{
    \sum_j |\eta_j|(\lambda_{\mathbf{z}^{(j)}_i}^\kappa-1)
    } \right)\right)_{i=1}^{|\mathcal{V}|}.
    \label{eq:mobius-gyromidpoint-residual}
\end{align}
This operation ensures the point set constraint of $\mathbb{D}$ for the residual flow. We recover the arithmetic mean as $\kappa\to 0$. During diffusion, Eq.~(\ref{eq:residual-diffusion-hyperbolic}) retains at least a portion of the initial and prior embeddings. Since the initial embedding possesses high energy, the residual connection mitigates energy degradation and retains the energy of the final iteration at the same level as the preceding iterations.

%% file: Content/Experiment.tex
\section{Empirical Results}
\subsection{Experiment Setup}
\textbf{Datasets} Under homophilic setting, we consider $5$ datasets for node classification and link prediction: \textsc{Disease}, \textsc{Airport} (transductive datasets, provided in \cite{chami2019hyperbolic} to investigate the tree-likeness modeling), \textsc{PubMed}, \textsc{CiteSeer} and \textsc{Cora} (\cite{yang2016revisiting} widely used citation networks), which are summarized in the table in Appendix~{A}. Additionally, we report the Gromov's hyperbolicity $\delta$ given by \cite{gromov1987hyperbolic} for each dataset. A graph is more hyperbolic as $\delta\to 0$ and is a tree when $\delta = 0$. 

For heterophilic datasets, we evaluate node classification on three heterophilic graphs, respectively, \textsc{Cornell}, \textsc{Texas} and \textsc{Wisconsin} \cite{pei2020geom} from the WebKB dataset (webpage networks). Detailed statistics are summarized in Appendix~{A}. We use the original fixed 10 split datasets. In addition, we report the homophily level $\mathcal{H}$ of each dataset, a sufficiently low $\mathcal{H}\le 0.3$ means that the dataset is more heterophilic when most of neighbours are not in the same class.

\textbf{Baselines} We compare our models to (1) \textit{Euclidean-hyperbolic} baselines, (2) \textit{discrete-continuous depth} baselines and (3) \textit{heterophilic relationship baselines}. For (1), we compare against feature-based models, Euclidean, and hyperbolic graph-based models. Feature-based models: without using graph structure, we feed node feature directly to MLP and HNN \cite{ganea2018hyperbolic}; Euclidean graph-based models: GCN \cite{kipf2016semi}, GAT \cite{velivckovic2017graph}, GraphSAGE \cite{hamilton2017inductive}, and SGC \cite{wu2019simplifying}; Hyperbolic graph-based models: HGCN \cite{chami2019hyperbolic}, $\kappa$GCN \cite{bachmann2020constant}, LGCN \cite{zhang2021lorentzian} and HyboNet \cite{chen2021fully}. For (2), we compare our models on citation networks with the discrete-continuous depth models. Discrete depth: GCNII \cite{chen2020simple}, C-DropEdge \cite{huang2021towards}; Discrete-decouple: HyLa-SGC \cite{yu2022random}; Continuous depth: GDE \cite{poli2019graph}, GRAND and BLEND \cite{chamberlain2021grand,chamberlain2021beltrami}. For (3), we compare to the prevalent GNNs: GCN, GAT, HGCN, HyboNet, and those optimized for heterophilic relationships: H2GCN \cite{zhu2020beyond}, GCNII, GraphSAGE and GraphCON \cite{rusch2022graph}. The test results are partially derived from the above works. For fairness, we compare to models with no more than $16$ layers/iterations. Please refer to Appendix~A for more details regarding the compared baselines. We detail the parameter settings for model and evaluation metric in Appendix~C.

\begin{table}[t]
    \begin{minipage}{.49\linewidth}
      \caption{Test accuracy (\%) for node classification task.}
      \centering
        \resizebox{0.99\linewidth}{!}{\input{table/nc_only}}
        \label{tb:nc-only}
    \end{minipage}%
    \hfill
    \begin{minipage}{.49\linewidth}
      \centering
        \caption{Test ROC AUC (\%) results for link prediction task.}
        \resizebox{0.99\linewidth}{!}{\input{table/lp_only}}
        \label{tb:lp-only}
    \end{minipage} 
\end{table}

\begin{table}[t]
    \begin{minipage}{.46\linewidth}
      \caption{Discrete-continuous depth GNN comparison.}
      \centering
        \resizebox{0.99\linewidth}{!}{\input{table/cmp_deep}}
        \label{tb:cmp-deep}
    \end{minipage}%
    \hfill
    \begin{minipage}{.52\linewidth}
      \centering
        \caption{Heterophilic relationship GNN comparison.}
        \resizebox{0.99\linewidth}{!}{\input{table/hetero}}
        \label{tb:cmp-hetero}
    \end{minipage} 
\end{table}

\subsection{Experiment Results}

\textbf{Euclidean-Hyperbolic Baselines} We investigate our methods with different solvers with $\tau=1$, \textit{i.e.} HGDE-E (multi-step explicit integrator, HRK4) and HGDE-I (multi-step implicit integrator, HAM). The experimental results are summarized in Tab.~\ref{tb:nc-only}-\ref{tb:lp-only}. (1) Our proposed models outperform previous Euclidean and hyperbolic models in four out of five datasets, suggesting that graph learning in hyperbolic space through topological diffusion is beneficial. (2) Hyperbolic models typically exhibit poor performance on datasets that are less hyperbolic (\textit{e.g.,} \textsc{Cora}), while our method surprisingly exceeds Euclidean GAT on datasets with lower $\delta$, indicating the necessity of curvature diffusion in adapting to datasets with scarce hierarchical structures and modeling long-term dependency via the local-global diffusivity function. (3) HGDE and other hyperbolic models achieve superior performance compared to Euclidean counterparts in link prediction due to the larger embedding space in hyperbolic geometry, which better preserves structural dependencies and allows for improved node arrangement. (4) HGDE-E generally outperforms HGDE-I with lower memory consumption and better precision, indicating that a larger $\tau$ may be necessary for implicit solvers. To align with multi-layer GNN schema (step-size is analogous to depth), we employ HGDE with HRK4 ($\tau=1$) for further evaluation.

\textbf{Discrete-Continuous Depth Baselines} In Tab.~\ref{tb:cmp-deep}, we compare our models with discrete and continuous-depth baselines. We observe that our method with $T\in[12, 16]$ achieves competitive results with the state-of-the-art models. Notably, HGDE models consistently outperform discrete models and continuous models with Euclidean embeddings, highlighting the benefits of utilizing hyperbolic embeddings in a continuous-depth framework. Compared to position encoding approaches (\textit{e.g.,} HyLa, BLEND), HGDE exhibits superior performance, indicating the feasibility of using hyperbolic space embeddings directly over initial position encoding. Interestingly, we find HGDE models performs better when increasing $T$ up to $12$, but slightly worse at $T=16$. This may due to the capacity of Poincar\'e ball or potential over-smoothing. Overall, the results underscore the effectiveness of the proposed HGDE models in harnessing the power of hyperbolic space for graph data modeling.

\begin{table}[t]
    \caption{Evaluation on image (\textsc{CIFAR}/\textsc{STL}) and text (\textsc{20News}) classification (Left) and Memory \& Runtime comparison (Right). $\star$ indicate OOM.}
    \begin{minipage}{.745\linewidth}
      \centering
        \resizebox{1.0\linewidth}{!}{\input{table/imagetext}}
    \end{minipage}%
    \hfill
    \begin{minipage}{.245\linewidth}
      \centering
        \resizebox{1.0\linewidth}{!}{\input{table/memoryc}}
    \end{minipage} 
    \label{tb:image-text-cls}
\end{table}

\textbf{Heterophilic Relationship Baselines} We show that HGDE is also capable in managing heterophilic relationship. In Tab.~\ref{tb:cmp-hetero}, HGDE achieves the highest scores on the \textsc{Texas} and \textsc{Cornell} and a competitive score on \textsc{Wisconsin}. This shows that hyperbolic space is beneficial in learning hierarchical heterophilic relationships. It also reflects the flexibility of HGDE as a hyperbolic vector flow for embedding high-order structures, with our model, powered by HPDE, outperforming other baselines on average.


\textbf{Image and Text Classification} We follow the experiment setup in \cite{wu2023difformer} and conduct additional experiments on the \textsc{CIFAR}, \textsc{STL}, and \textsc{20News} datasets to evaluate HGDE in multiple scenarios with limited label rates. We employ the SimCLR \cite{chen2020simclr} extracted embedding as provided in \cite{wu2023difformer} for image classification. For the pre-processed \textsc{20News} \cite{pedregosa2011scikit} for text classification, we take $10$ topics and regard words with $\text{TF-IDF} > 5$ as features. For graph-based models, we use \textit{k}NN to construct a graph over input features. For HGDE (hyperbolic), we map the initial feature to $\mathbb{D}_\kappa$ via $\exp_\mathbf{o}^\kappa(\cdot)$ before the embedding process. As depicted in Tab.~\ref{tb:image-text-cls}(Left), HGDE consistently surpasses its opponents, including MLP, LabelProp \cite{zhu2003semi}, ManiReg \cite{belkin2006manifold}, GCN-\textit{k}NN, GAT-\textit{k}NN, DenseGAT, and GLCN \cite{jiang2019semi}. Across all datasets, HGDE outperforms the Euclidean models, underscoring its proficiency in understanding the potential hierarchical structure of image embeddings \cite{khrulkov2020hyperbolic} and text embeddings. Furthermore, HGDE exhibits good performance compared to static graph-based baselines \textit{e.g.,} GAT-\textit{k}NN and GLCN, which underlines the distinct advantage of the evolving diffusivity mechanism in understanding the potential hierarchical structure of image/text embeddings.

\subsection{Ablation Study}

\begin{figure}[t]
    \centering
    \begin{minipage}[b]{0.6\linewidth}
        \begin{minipage}[b]{\linewidth}
            \centering
            \begin{subfigure}[b]{0.49\linewidth}
                \centering
                \includegraphics[width=\linewidth]{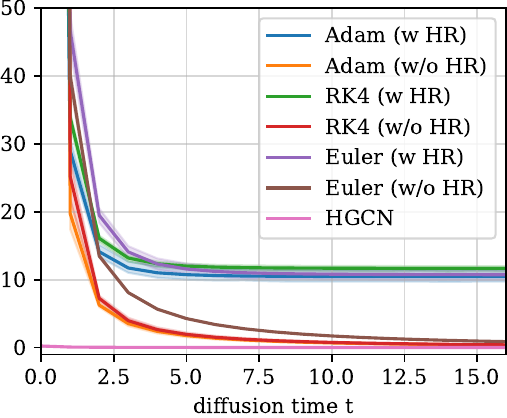}
                \label{Fig.sub.cora-energy}
            \end{subfigure}
            \hfill
            \begin{subfigure}[b]{0.49\linewidth}
                \centering
                \includegraphics[width=\linewidth]{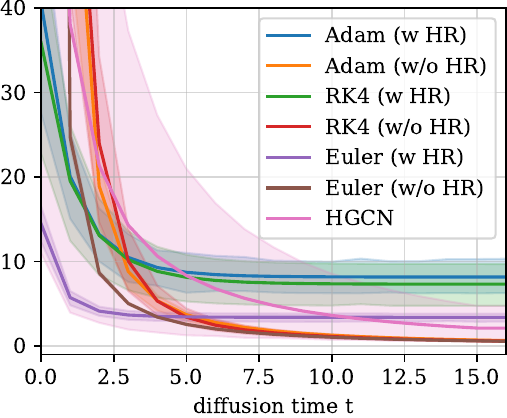}
                \label{Fig.sub.citeseer-energy}
            \end{subfigure}
            \caption{Hyperbolic Dirichlet energy $f^{\kappa}_\mathrm{DE}(\cdot)$ variation through $t$ on \textsc{Cora} (left) and \textsc{CiteSeer} (right). Models are compared with different integrators w or w/o hyperbolic residual.}
            \label{Fig.energy-convergence-ablation}
        \end{minipage}
        \vfill
        \begin{minipage}[b]{\linewidth}
            \begin{subfigure}[b]{\linewidth}
                \centering
                \includegraphics[width=\linewidth]{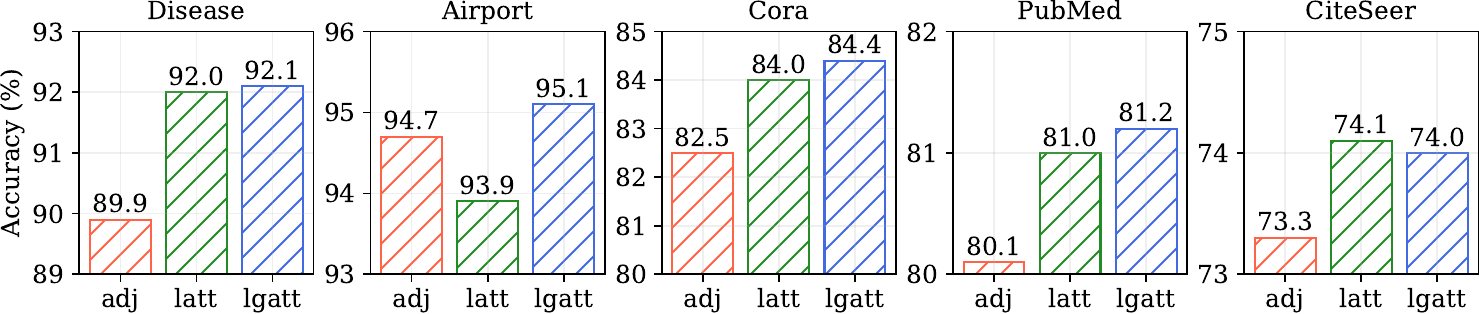}
                \label{Fig.sub.cmpatt}
            \end{subfigure}
            \caption{Averaged node classification performance comparison of models with different diffusivity functions on various datasets.}
        \end{minipage}
    \end{minipage}
    \hfill
    \begin{minipage}[b]{0.38\linewidth}
        \centering
        \begin{subfigure}[b]{0.99\linewidth}
            \centering
            \fbox{\includegraphics[width=\linewidth]{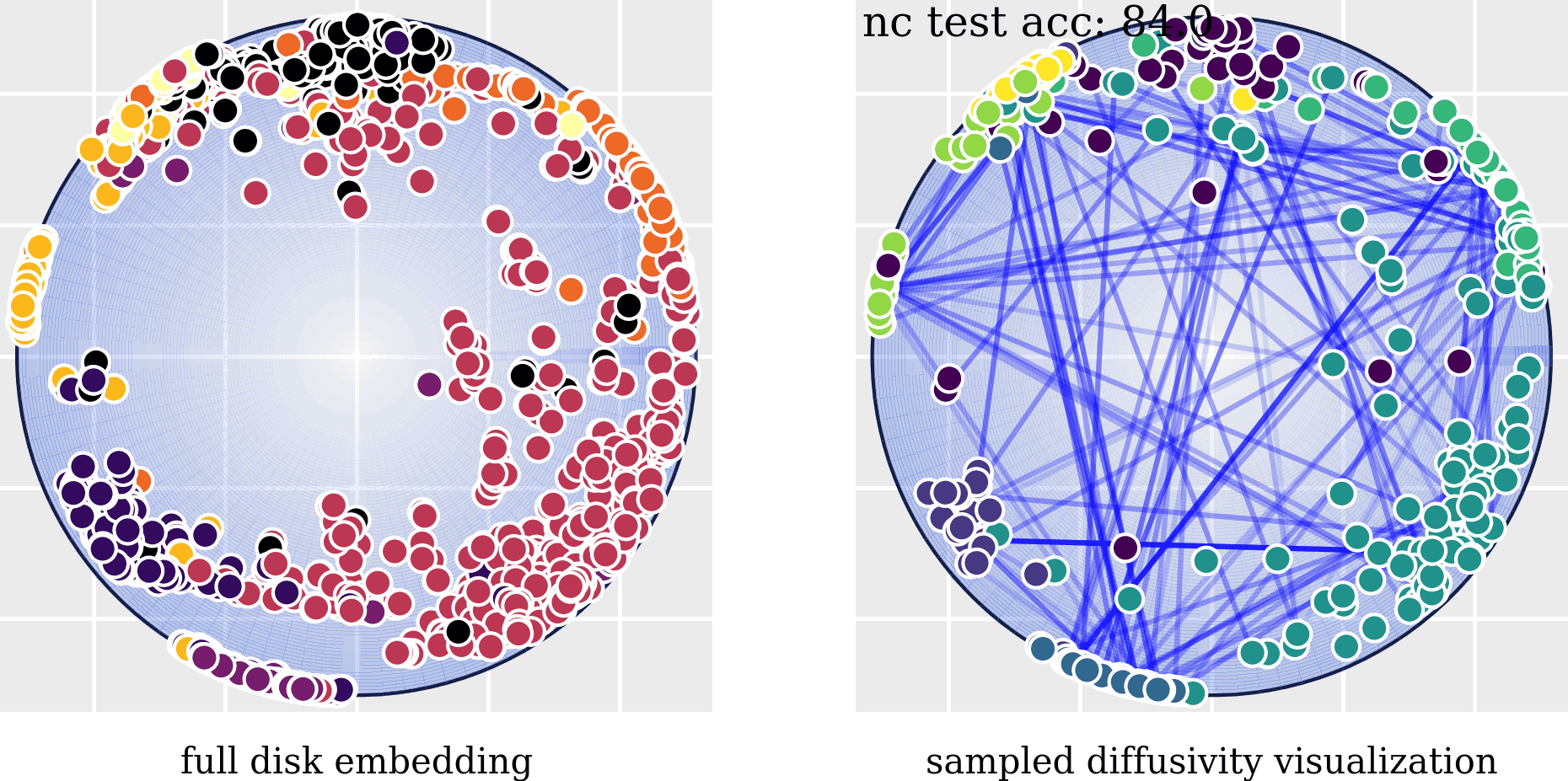}}
            \label{Fig.sub.latt}
        \end{subfigure}
        \vfill
        \begin{subfigure}[b]{0.99\linewidth}
            \centering
            \fbox{\includegraphics[width=\linewidth]{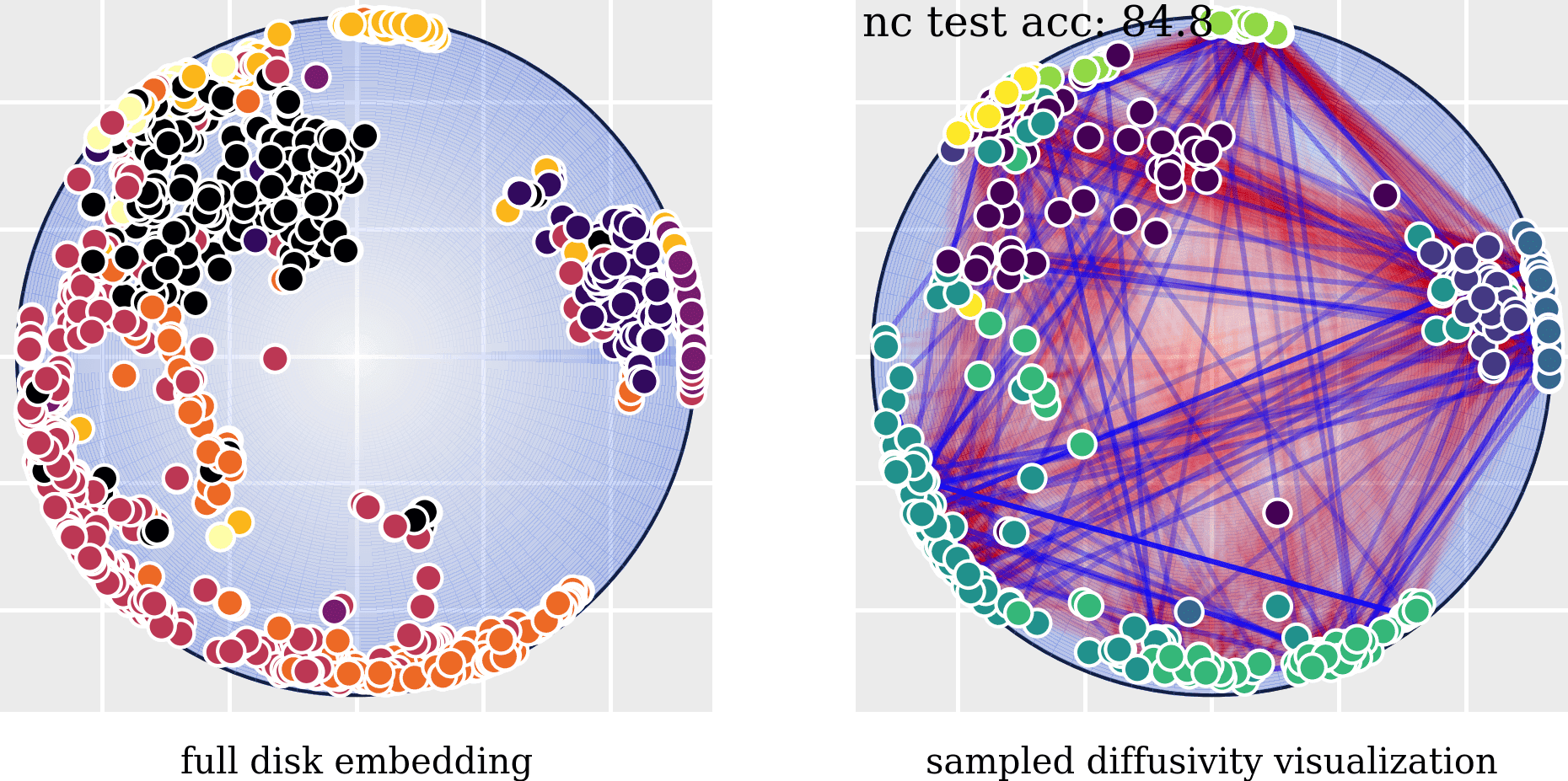}}
            \label{Fig.sub.lgatt}
        \end{subfigure}
        \caption{\textsc{Cora} diffusivity (400 node sampled from $\mathbb{D}^2_{\kappa}$ embeddings) produced by $a^{\mathrm{ldiff}}$ (left) and $a^{\mathrm{lgdiff}}$ (right), blue and red lines denote local and global attention; bolder lines indicate more attentiveness.}
        \label{Fig.att-weight-vis}
    \end{minipage}
\end{figure}

\textbf{Efficacy of Hyperbolic Residual} Fig.~\ref{Fig.energy-convergence-ablation} visualizes the convergence of hyperbolic energy through iterations. We observe that, without residuals, the averaged energy rapidly decreases to near-zero values, supporting the hypothesis that, without residual connections, the embedding can evolve to an overly smoothed state that is potentially low in expressiveness. However, with hyperbolic residuals, for all three integrators, the average energy decreases over the first few iterations and then appears to stabilize around a certain value above zero. This behavior is consistent across both datasets, suggesting that the system is able to converge to a stable state with non-zero energy.


\textbf{Efficacy of Diffusivity Function} Fig.~\ref{Fig.att-weight-vis} visualizes sampled node embeddings and their edge diffusivity on \textsc{Cora}. The blue edges are inherently determined by the graph structure. Red ones are determined by global attention, showing that $a^\mathrm{lgdiff}$ accounts for high-order relations. The bar graph shows the average accuracy on various datasets produced by HGDE with different diffusivity functions. We find that anisotropic approaches generally outperform the isotropic approach, suggesting the necessity of directional information in the diffusion process. Although the performance degrades on \textsc{CiteSeer} when using $a^{\mathrm{lgdiff}}$, there are significant improvements on other graphs, certifying the benefit of higher-order proximity induced by local-global diffusivity.

\textbf{Parameter Efficiency} In Tab.~\ref{tb:image-text-cls}~(Right), we provide an additional comparison of peak GPU memory usage and per-epoch running time on the \textsc{Cora}. We tested HGDE-E where all models have a $16$ hidden dim. Our model significantly outperforms the other baselines in both training time (for $\ge 4$ layers) and memory consumption. The memory reduction is primarily due to the utilization of sparse attention, and the advantages of a weight-tied network (requiring only single-layer parameters) as a nature of HPDE. The training time efficiency is achieved by eliminating layer-wise feature transformation, implementing weight-tying, and applying scattered-agg for hyperbolic representation.

%% file: table/nc_only.tex
\begin{NiceTabular}{@{}llllll@{}}[colortbl-like]
\toprule
Dataset     & \textsc{Disease}                                 & \textsc{Airport}                                 & \textsc{PubMed}                                  & \textsc{CiteSeer}                                & \textsc{Cora}                                    \\
$\delta$    & $0$                                              & $1$                                              & $3.5$                                            & $5$                                              & $11$                                             \\ \midrule
MLP         & 32.5$_{\pm 1.1}$                                 & 60.9$_{\pm 3.4}$                                 & 72.4$_{\pm 0.2}$                                 & 59.5$_{\pm 0.9}$                                 & 51.6$_{\pm 1.3}$                                 \\
HNN         & 45.5$_{\pm 3.3}$                                 & 80.6$_{\pm 0.5}$                                 & 69.9$_{\pm 0.4}$                                 & 59.5$_{\pm 1.2}$                                 & 54.7$_{\pm 0.6}$                                 \\
GCN         & 69.7$_{\pm 0.4}$                                 & 81.6$_{\pm 0.6}$                                 & 78.1$_{\pm 0.4}$                                 & 70.3$_{\pm 0.4}$                                 & 81.5$_{\pm 0.5}$                                 \\
GAT         & 70.4$_{\pm 0.4}$                                 & 82.7$_{\pm 0.4}$                                 & 78.2$_{\pm 0.4}$                                 & {\color[HTML]{6200C9} \textbf{71.6$_{\pm 0.8}$}}                        & 83.0$_{\pm 0.5}$                                 \\
SAGE        & 69.1$_{\pm 0.6}$                                 & 82.2$_{\pm 0.5}$                                 & 77.5$_{\pm 2.4}$                                 & 67.5$_{\pm 0.7}$                                 & 79.9$_{\pm 2.5}$                                 \\
SGC         & 69.5$_{\pm 0.2}$                                 & 80.6$_{\pm 0.2}$                                 & 78.8$_{\pm 0.2}$                                 & 71.4$_{\pm 0.8}$                                 & 81.3$_{\pm 0.5}$                                 \\ \midrule
HGCN        & 82.8$_{\pm 0.8}$                                 & 89.2$_{\pm 1.3}$                                 & {\color[HTML]{6200C9} \textbf{80.3$_{\pm 0.3}$}}                        & 68.0$_{\pm 0.6}$                                 & 79.9$_{\pm 0.2}$                                 \\
$\kappa$GCN & 82.1$_{\pm 1.1}$                                 & 84.4$_{\pm 0.4}$                                 & 78.3$_{\pm 0.6}$                                 & 71.1$_{\pm 0.6}$                                 & 80.8$_{\pm 0.6}$                                 \\
LGCN        & 84.4$_{\pm 0.8}$                                 & {\color[HTML]{6200C9} \textbf{90.9$_{\pm 1.0}$}}                        & 78.8$_{\pm 0.5}$                                 & 71.1$_{\pm 0.3}$                                 & {\color[HTML]{6200C9} \textbf{83.3$_{\pm 0.5}$}}                        \\
HyboNet     & {\color[HTML]{FE0000} \textbf{96.0$_{\pm 1.0}$}} & 90.9$_{\pm 1.4}$                                 & 78.0$_{\pm 1.0}$                                 & 69.8$_{\pm 0.6}$                                 & 80.2$_{\pm 1.3}$                                 \\ \midrule
\rowcolor{Gray!30}
HGDE-E      & {\color[HTML]{3531FF} \textbf{92.1$_{\pm 1.6}$}} & {\color[HTML]{FE0000} \textbf{95.1$_{\pm 0.4}$}} & {\color[HTML]{FE0000} \textbf{81.2$_{\pm 0.5}$}} & {\color[HTML]{FE0000} \textbf{74.1$_{\pm 0.5}$}} & {\color[HTML]{FE0000} \textbf{84.4$_{\pm 0.7}$}} \\
\rowcolor{Gray!30}
HGDE-I      & {\color[HTML]{6200C9} \textbf{90.9$_{\pm 2.5}$}}                        & {\color[HTML]{3531FF} \textbf{93.9$_{\pm 0.8}$}} & {\color[HTML]{3531FF} \textbf{81.0$_{\pm 0.3}$}} & {\color[HTML]{3531FF} \textbf{73.5$_{\pm 0.7}$}} & {\color[HTML]{3531FF} \textbf{84.0$_{\pm 0.4}$}} \\ \bottomrule
\end{NiceTabular}

%% file: table/lp_only.tex
\begin{NiceTabular}{@{}llllll@{}}[colortbl-like]
\toprule
Dataset     & \textsc{Disease}                                 & \textsc{Airport}                                 & \textsc{PubMed}                                  & \textsc{CiteSeer}                                & \textsc{Cora}                                    \\
$\delta$    & $0$                                              & $1$                                              & $3.5$                                            & $5$                                              & $11$                                             \\ \midrule
MLP         & 69.9$_{\pm 3.4}$                                 & 68.9$_{\pm 0.5}$                                 & 83.3$_{\pm 0.6}$                                 & 93.7$_{\pm 0.6}$                                 & 83.3$_{\pm 0.6}$                                 \\
HNN         & 70.2$_{\pm 0.1}$                                 & 80.6$_{\pm 0.5}$                                 & 94.7$_{\pm 0.1}$                                 & 93.3$_{\pm 0.5}$                                 & 90.9$_{\pm 0.4}$                                 \\
GCN         & 64.7$_{\pm 0.5}$                                 & 89.3$_{\pm 0.4}$                                 & 89.6$_{\pm 3.7}$                                 & 82.6$_{\pm 1.9}$                                 & 90.5$_{\pm 0.2}$                                 \\
GAT         & 69.8$_{\pm 0.3}$                                 & 90.9$_{\pm 0.2}$                                 & 91.5$_{\pm 1.8}$                                 & 86.5$_{\pm 1.5}$                                 & 93.2$_{\pm 0.2}$                                 \\
SAGE        & 65.9$_{\pm 0.3}$                                 & 90.4$_{\pm 0.5}$                                 & 86.2$_{\pm 0.8}$                                 & 92.1$_{\pm 0.4}$                                 & 85.5$_{\pm 0.5}$                                 \\
SGC         & 65.1$_{\pm 0.2}$                                 & 89.8$_{\pm 0.3}$                                 & 94.1$_{\pm 0.1}$                                 & 91.4$_{\pm 1.7}$                                 & 91.5$_{\pm 0.2}$                                 \\ \midrule
HGCN        & 91.2$_{\pm 0.6}$                                 & 96.4$_{\pm 0.1}$                                 & 95.1$_{\pm 0.1}$                                 & {\color[HTML]{6200C9} \textbf{96.6$_{\pm 0.1}$}}                        & {\color[HTML]{6200C9} \textbf{93.8$_{\pm 0.1}$}}                        \\
$\kappa$GCN & 92.0$_{\pm 0.5}$                                 & 92.5$_{\pm 0.5}$                                 & 94.9$_{\pm 0.3}$                                 & 95.1$_{\pm 0.6}$                                 & 92.6$_{\pm 0.4}$                                 \\
LGCN        & {\color[HTML]{3531FF} \textbf{96.6$_{\pm 0.6}$}} & 96.0$_{\pm 0.6}$                                 & {\color[HTML]{FE0000} \textbf{96.6$_{\pm 0.1}$}} & 95.8$_{\pm 0.4}$                                 & 93.6$_{\pm 0.4}$                                 \\
HyboNet     & {\color[HTML]{FE0000} \textbf{96.8$_{\pm 0.4}$}} & {\color[HTML]{6200C9} \textbf{97.3$_{\pm 0.3}$}}                        & 95.8$_{\pm 0.2}$                                 & {\color[HTML]{3531FF} \textbf{96.7$_{\pm 0.8}$}} & 93.6$_{\pm 0.3}$                                 \\ \midrule
\rowcolor{Gray!30}
HGDE-E      & {\color[HTML]{6200C9} \textbf{96.2$_{\pm 0.5}$}}                        & {\color[HTML]{FE0000} \textbf{98.2$_{\pm 0.2}$}} & {\color[HTML]{3531FF} \textbf{96.6$_{\pm 0.2}$}} & {\color[HTML]{FE0000} \textbf{96.7$_{\pm 0.7}$}} & {\color[HTML]{3531FF} \textbf{94.1$_{\pm 0.4}$}} \\
\rowcolor{Gray!30}
HGDE-I      & 95.6$_{\pm 0.5}$                                 & {\color[HTML]{3531FF} \textbf{97.6$_{\pm 0.5}$}} & {\color[HTML]{6200C9} \textbf{96.2$_{\pm 0.7}$}}                        & 96.4$_{\pm 0.7}$                                 & {\color[HTML]{FE0000} \textbf{94.5$_{\pm 0.8}$}} \\ \bottomrule
\end{NiceTabular}

%% file: table/cmp_deep.tex
\begin{NiceTabular}{@{}c|llll@{}}[colortbl-like]
\toprule
Type                                                                               & Model      & \textsc{Cora}                                    & \textsc{CiteSeer}                                & \textsc{PubMed}                                  \\ \midrule
                                                                                   & GCNII      & {\color[HTML]{FE0000} \textbf{84.6$_{\pm 0.8}$}} & 72.9$_{\pm 0.5}$                                 & 80.2$_{\pm 0.4}$                                 \\
\multirow{-2}{*}{Discrete}                                                         & C-DropEdge & 82.6$_{\pm 0.9}$                                 & 71.0$_{\pm 1.0}$                                 & 77.8$_{\pm 1.0}$                                 \\ \midrule
\begin{tabular}[c]{@{}c@{}}Decouple\\ (Hyp PosEnc)\end{tabular}                    & HyLa-SGC   & 82.5$_{\pm 0.5}$                                 & 72.6$_{\pm 1.0}$                                 & 80.3$_{\pm 0.9}$                                 \\ \midrule
                                                                                   & GDE        & 83.8$_{\pm 0.5}$                                 & 72.5$_{\pm 0.5}$                                 & 79.9$_{\pm 0.3}$                                 \\
\multirow{-2}{*}{Continuous}                                                       & GRAND      & 82.9$_{\pm 0.7}$                                 & 73.6$_{\pm 0.3}$                                 & {\color[HTML]{3531FF} \textbf{81.0$_{\pm 0.4}$}} \\ \midrule
\begin{tabular}[c]{@{}c@{}}Continuous\\ (Hyp PosEnc)\end{tabular}                  & BLEND      & {\color[HTML]{6200C9} \textbf{84.2$_{\pm 0.6}$}} & {\color[HTML]{FE0000} \textbf{74.4$_{\pm 0.7}$}} & 80.7$_{\pm 0.7}$                                 \\ \midrule
\rowcolor{Gray!30}
                                                                                   & HGDE(4)    & 83.4$_{\pm 0.5}$                                 & 73.0$_{\pm 0.3}$                                 & 80.2$_{\pm 0.6}$                                 \\
\rowcolor{Gray!30}
                                                                                   & HGDE(8)    & 83.7$_{\pm 0.6}$                                 & 73.5$_{\pm 0.7}$                                 & 80.8$_{\pm 0.4}$                                 \\
\rowcolor{Gray!30}
                                                                                   & HGDE(12)   & {\color[HTML]{6200C9} \textbf{84.2$_{\pm 0.6}$}} & {\color[HTML]{3531FF} \textbf{74.1$_{\pm 0.5}$}} & {\color[HTML]{FE0000} \textbf{81.2$_{\pm 0.5}$}} \\
\rowcolor{Gray!30}
\multirow{-4}{*}{\begin{tabular}[c]{@{}c@{}}Continuous\\ (Hyp Embed)\end{tabular}} & HGDE(16)   & {\color[HTML]{3531FF} \textbf{84.4$_{\pm 0.7}$}} & {\color[HTML]{6200C9} \textbf{73.8$_{\pm 0.7}$}} & {\color[HTML]{6200C9} \textbf{80.9$_{\pm 0.3}$}} \\ \bottomrule
\end{NiceTabular}

%% file: table/hetero.tex
\begin{NiceTabular}{@{}c|llll@{}}[colortbl-like]
\toprule
                                                                            &               & \textsc{Texas}                                   & \textsc{Wisconsin}                               & \textsc{Cornell}                                 \\
\multirow{-2}{*}{Type}                                                      & $\mathcal{H}$ & 0.11                                             & 0.21                                             & 0.30                                             \\ \midrule
                                                                            & GCN           & 55.1$_{\pm 5.2}$                                 & 51.8$_{\pm 3.1}$                                 & 60.5$_{\pm 5.3}$                                 \\
\multirow{-2}{*}{Euclidean}                                                 & GAT           & 52.2$_{\pm 6.6}$                                 & 49.4$_{\pm 4.1}$                                 & 61.9$_{\pm 5.1}$                                 \\ \midrule
                                                                            & HGCN          & 55.7$_{\pm 6.3}$                                 & 48.1$_{\pm 6.1}$                                 & 62.1$_{\pm 3.7}$                                 \\
\multirow{-2}{*}{Hyperbolic}                                                & HyboNet       & 60.0$_{\pm 4.1}$                                 & 51.2$_{\pm 3.3}$                                 & 62.3$_{\pm 3.5}$                                 \\ \midrule
                                                                            & H2GCN         & {\color[HTML]{3531FF} \textbf{84.9$_{\pm 7.2}$}} & {\color[HTML]{6200C9} \textbf{87.7$_{\pm 5.0}$}} & {\color[HTML]{3531FF} \textbf{82.7$_{\pm 5.3}$}} \\
                                                                            & GCNII         & 77.6$_{\pm 3.8}$                                 & 80.4$_{\pm 3.4}$                                 & 77.9$_{\pm 3.8}$                                 \\
                                                                            & SAGE          & 82.4$_{\pm 6.1}$                                 & 81.2$_{\pm 5.6}$                                 & 76.0$_{\pm 5.0}$                                 \\
\multirow{-4}{*}{\begin{tabular}[c]{@{}c@{}}High-Order\\ GNNs\end{tabular}} & GraphCON      & {\color[HTML]{6200C9} \textbf{85.4$_{\pm 4.2}$}} & {\color[HTML]{FE0000} \textbf{87.8$_{\pm 3.3}$}} & {\color[HTML]{6200C9} \textbf{84.3$_{\pm 4.8}$}} \\ \midrule
\rowcolor{Gray!30}
Ours                                                                        & HGDE          & {\color[HTML]{FE0000} \textbf{85.9$_{\pm 2.8}$}} & {\color[HTML]{3531FF} \textbf{86.2$_{\pm 2.4}$}} & {\color[HTML]{FE0000} \textbf{85.0$_{\pm 5.3}$}} \\ \bottomrule
\end{NiceTabular}

%% file: table/imagetext.tex
\setlength{\aboverulesep}{0pt}
\setlength{\belowrulesep}{0pt}
\setlength{\extrarowheight}{.75ex}
\begin{NiceTabular}{@{}cc|ccccccc|>{\columncolor{Gray!30}}c@{}}[colortbl-like]
\toprule
\multicolumn{2}{c|}{\textbf{Dataset}} & MLP & LabelProp & ManiReg & GCN-\textit{k}NN & GAT-\textit{k}NN & DenseGAT & GLCN & HGDE \\ 
\midrule
\multicolumn{1}{c|}{\multirow{3}{*}{\textsc{CIFAR}}} & 100 labels & 65.9$_{\pm 1.3}$ & 66.2$_{}$ & 67.0$_{\pm 1.9}$ & 66.7$_{\pm 1.5}$ & 66.0$_{\pm 2.1}$ & $\star$ & 66.6$_{\pm 1.4}$ & {\textbf{68.9}$_{\pm 2.1}$} \\
\multicolumn{1}{c|}{} & 500 labels & 73.2$_{\pm 0.4}$ & 70.6$_{}$ & 72.6$_{1\pm .2}$ & 72.9$_{\pm 0.4}$ & 72.4$_{\pm 0.5}$ & $\star$ & 72.7$_{\pm 0.5}$ & \textbf{74.0}$_{\pm 1.8}$ \\
\multicolumn{1}{c|}{} & 1000 labels & 75.4$_{\pm 0.6}$ & 71.9$_{}$ & 74.3$_{\pm 0.4}$ & 74.7$_{\pm 0.5}$ & 74.1$_{\pm 0.5}$ & $\star$ & 74.7$_{\pm 0.3}$ & \textbf{76.3}$_{\pm 0.9}$ \\ 
\midrule
\multicolumn{1}{c|}{\multirow{3}{*}{\textsc{STL}}} & 100 labels & 66.2$_{\pm 1.4}$ & 65.2$_{}$ & 66.5$_{\pm 1.9}$ & 66.9$_{\pm 0.5}$ & 66.5$_{\pm 0.8}$ & $\star$ & 66.4$_{\pm 0.8}$ & \textbf{66.9}$_{\pm 1.3}$ \\
\multicolumn{1}{c|}{} & 500 labels & \textbf{73.0}$_{\pm 0.8}$ & 71.8$_{}$ & 72.5$_{\pm 0.5}$ & 72.1$_{\pm 0.8}$ & 72.0$_{\pm 0.8}$ & $\star$ & 72.4$_{\pm 1.3}$ & 72.5$_{\pm 0.2}$ \\
\multicolumn{1}{c|}{} & 1000 labels & 75.0$_{\pm 0.8}$ & 72.7$_{}$ & 74.2$_{\pm 0.5}$ & 73.7$_{\pm 0.4}$ & 73.9$_{\pm 0.6}$ & $\star$ & 74.3$_{\pm 0.7}$ & \textbf{75.1}$_{\pm 0.6}$ \\ 
\midrule
\multicolumn{1}{c|}{\multirow{3}{*}{\textsc{20News}}} & 1000 labels & 54.1$_{\pm 0.9}$ & 55.9$_{}$ & 56.3$_{\pm 1.2}$ & 56.1$_{\pm 0.6}$ & 55.2$_{\pm 0.8}$ & 54.6$_{\pm 0.2}$ & 56.2$_{\pm 0.8}$ & \textbf{56.3}$_{\pm 0.9}$ \\
\multicolumn{1}{c|}{} & 2000 labels & 57.8$_{\pm 0.9}$ & 57.6$_{}$ & 60.0$_{\pm 0.8}$ & 60.6$_{\pm 1.3}$ & 59.1$_{\pm 2.2}$ & 59.3$_{\pm 1.4}$ & 60.2$_{\pm 0.7}$ & \textbf{61.0}$_{\pm 1.0}$ \\
\multicolumn{1}{c|}{} & 4000 labels & 62.4$_{\pm 0.6}$ & 59.5$_{}$ & 63.6$_{\pm 0.7}$ & 64.3$_{\pm 1.0}$ & 62.9$_{\pm 0.7}$ & 62.4$_{\pm 1.0}$ & 64.1$_{\pm 0.8}$ & \textbf{64.1}$_{\pm 0.8}$ \\ 
\bottomrule
\end{NiceTabular}

%% file: table/memoryc.tex
\begin{NiceTabular}{@{}c|lll@{}}[colortbl-like]
\toprule
$T/\tau$  & Model (with Att) & Memory ($\times 10^6$) & Runtime (ms)   \\ \midrule
\multirow{5}{*}{2}  & HGCN             & 4045                   & \textbf{9.25}  \\
                    & HGCN (LocalAtt)  & 4246                   & 1310.31        \\
                    & LGCN             & 4630                   & 16.64          \\
                    & HyboNet          & 4368                   & 14.90          \\
\rowcolor{Gray!30}
\cellcolor{white}  & HGDE             & \textbf{62}            & 13.66          \\ \midrule
\multirow{5}{*}{4}  & HGCN             & 10255                  & 29.77          \\
                    & HGCN (LocalAtt)  & 10578                  & 4086.08        \\
                    & LGCN             & 12675                  & 40.88          \\
                    & HyboNet          & 11931                  & 35.23          \\
\rowcolor{Gray!30}
\cellcolor{white} & HGDE             & \textbf{73}            & \textbf{20.28} \\ \midrule
\multirow{5}{*}{8}  & HGCN             & 22674                  & 67.24          \\
                    & HGCN (LocalAtt)  & OOM                    & $^\star$       \\
                    & LGCN             & 23712                  & 160.75         \\
                    & HyboNet          & 23403                  & 122.55         \\
\rowcolor{Gray!30}
\cellcolor{white}  & HGDE             & \textbf{112}           & \textbf{34.11} \\ \midrule
\multirow{2}{*}{16} & All Baselines    & OOM                    & $^\star$       \\
\rowcolor{Gray!30}
\cellcolor{white}  & HGDE             & \textbf{188}           & \textbf{61.95} \\ \bottomrule
\end{NiceTabular}

%% file: Content/Conclusion.tex
\section{Conclusion}
We developed multiple numerical integrators for HPDE, and proposed the first hyperbolic continuous-time embedding diffusion framework -- HGDE. Being capable of capturing both low and high order proximity, HGDE outperforms both Euclidean and hyperbolic baselines on various datasets. The effectiveness of HGDE was further validated by the ablation studies on hyperbolic energy and diffusivity functions. The superiority of HGDE underscores the potential of developing PDE-based non-Euclidean models.

\subsubsection{Limitation} 
While HGDE presents strong performance in modeling graph data, hyperbolic spaces may not always be optimal, particularly for data without clear hierarchical structures. For instance, HGDE is difficult to beat natural Euclidean deep models (\textit{e.g.} GCNII) on the non-hierarchical $\textsc{Cora}$. Moreover, a higher memory complexity and lower training time only tells the efficiency rather than scalability of HGDE, since our models are evaluated with fixed number of parameters (which is natural for ODE-based models), increasing $T$ is not necessarily \textit{scaling up}. Future work include addressing these limitations and exploring the scalability and generalizability of HGDE in diverse real-world settings.

%% file: Content/Appendices/AppA.tex
\section*{Appendix A: Baselines \& Statistics}
\subsection*{Dataset Statistics}
\begin{table}[H]
\centering
\caption{Statistics of homophilic benchmark datasets.}
\resizebox{0.6\linewidth}{!}{\input{table/datasets.tex}}
\label{tb:summary-dataset}
\end{table}

\begin{table}[H]
\centering
\caption{Statistics of heterophilic benchmark datasets.}
\resizebox{0.65\linewidth}{!}{\input{table/dataset_hetro.tex}}
\label{tb:summary-dataset-hetro}
\end{table}

\subsection*{Baselines}
To assess the efficacy of HGDE in node classification and link prediction, we include a number of prominent GNN/HGNN variants, as baselines, summarized as follows:
\subsubsection*{Shallow Euclidean Embedding Models}
\begin{itemize}
    \item \textbf{GCN/GAT.} Given a normalized edge weight matrix $\Tilde{P}$, let $w_{i,j}$ be the $ij$-th element of $\Tilde{P}$ \textit{i.e.,} the weight of edge $(i,j)\in \mathcal{E}$, then $\forall i:\sum_{j}w_{i,j} = 1$. GCN/GAT learns the node embedding by propagating messages over the $\Tilde{P}$. The difference is that GCN leverages the renormalization trick to obtain a fixed augmented Laplacian for all message propagation, while GAT learns dynamic $w_{i,j}$ which is normalized using softmax before each propagation.
    \item \textbf{(Hyperbolic Positional Encoding) HyLa-SGC.} SGC simplifies the GCN by decoupling hidden weights and activation functions from aggregation layers. HyLa is a RFF-like \cite{tancik2020fourier} positional encoding that approximates an isometry-invariant kernel over hyperbolic space. HyLa can be plugged as an end-to-end encoder before SGC to learn the hyperbolic geometric priors of the node features, one advantage of HyLa-SGC is its training efficiency compared to HGCN. HyLa-SGC embeds graph nodes in Euclidean space.
\end{itemize}
\subsubsection*{Shallow Non-Euclidean Embedding Models}
\begin{itemize}
    \item \textbf{HGCN.} To ensure the transformed features satisfy the hyperbolic geometry, HGCN defines feature transformation and message aggregation by pushing forward nodes to tangent space then pull them back to the hyperboloid. 
    \item \textbf{LGCN.} Different from HGCN, LGCN defines an augmented linear transformation and message aggregation using the Lorentz centroid, which is an improved weighted mean operator that reduces distortion compared to local diffeomorphisms, such as tangent space operations. 
    \item \textbf{HyboNet.} Further, the HyboNet derives a linear transformation within the hyperboloid without the dependence on tangent spaces, namely, the Lorentz linear layer.
    \item \textbf{$\kappa$GCN.} Beyond hyperbolic space, $\kappa$GCN generalize embedding geometry to products of $\kappa$-Stereographic models, which includes Euclidean space, Poincar\'e ball, hyperboloid, sphere, and projective hypersphere.
\end{itemize}

\subsubsection*{Discrete-Depth Euclidean Embedding Models}
\begin{itemize}
    \item \textbf{GCNII/(C)DropEdge.} To build deeper GCNs that mitigates over-smoothing, (C)DropEdge randomly removes a certain number of edges from the input graph at each training epoch, which reduces the convergence speed of over-smoothing. GCNII extend the vanilla GCN model with two simple techniques at each layer: an initial connection to the input and an identity mapping to regularize the weight; both techniques are proved to help prevent over-smoothing. 
\end{itemize}

\subsubsection*{Continuous-Depth Euclidean Embedding Models}
\begin{itemize}
    \item \textbf{GDE.} GDE is the first work to adapt neural ODEs to graphs, where the input-output correlation is established through a continuum of GCN layers. Instead of layer-wise propagation of discrete GCN-like models, GDE treats graph convolution as an ODE function, enabling simulation of continuous layer propagation.
    \item \textbf{GRAND.} GRAND consider the \textit{heat} diffusion instead of graph convolution as the ODE function. The diffusion process describes a directional node embedding (heat) variation induced by neighboring positions; the strength of such variation is called diffusivity, which is modeled by multi-head attention. In such a way, the attentive aggregation is applied on \textit{difference of node pairs} rather than directly on node embeddings.
    \item \textbf{(Hyperbolic Positional Encoding) BLEND.} According to \cite{song2022robustness}, BLEND is essentially GRAND with positional encoding. The input of PDE is an joint Euclidean node feature and its positional encoding via hyperbolic metric. The PDE solves the joint diffusion process that captures hyperbolic information. BLEND does not change the architecture of PDE solver, thus remains a Euclidean embedding model.
\end{itemize}

%% file: table/datasets.tex
\begin{tabular}{lrrrrr} 
\toprule
Dataset  & \# Nodes & \# Edges & Classes & Features & $\delta$  \\ 
\midrule
\textsc{Disease}  & 1,044    & 1,043    & 2       & 1,000    & 0         \\
\textsc{Airport}  & 3,188    & 18,631   & 4       & 4        & 1         \\
\textsc{PubMed}   & 19,717   & 44,338   & 3       & 500      & 3.5       \\
\textsc{CiteSeer} & 3,327    & 4,732    & 6       & 3,703    & 5         \\
\textsc{Cora}     & 2,708    & 5,429    & 7       & 1,433    & 11        \\
\bottomrule
\end{tabular}

%% file: table/dataset_hetro.tex
\begin{tabular}{lrrrrr} 
\toprule
Dataset  & \# Nodes & \# Edges & Classes & Features & $\mathcal{H}$  \\ 
\midrule
\textsc{Texas}  & 183    & 295    & 5       & 1,703    & 0.11         \\
\textsc{Cornell}  & 182    & 295   & 5       & 1,703        & 0.21         \\
\textsc{Wisconsin}   & 251   & 499   & 5       & 1,703      & 0.30       \\
\bottomrule
\end{tabular}

%% file: Content/Appendices/AppB.tex
\section*{Appendix B: Hyperbolic Geometry Operations}
\subsection*{Poincar\'e Ball Model}
The Poincar\'e ball is defined as the Riemannian manifold $\mathbb{D}^n_\kappa = (\mathcal{D}^n_\kappa, g^{\mathbb{D}})$, with point set $\mathcal{D}^n_\kappa = \{\mathbf{x} \in \mathbb{R}^n : \|\mathbf{x}\| < -\frac{1}{\kappa}\}$ and Riemannian metric
\begin{equation}
    g^{\mathbb{D}}_\mathbf{x} = (\lambda_\mathbf{x}^\kappa)^2 \mathbf{I}_n,
\end{equation}
where $\lambda_\mathbf{x}^\kappa = \frac{2}{1+\kappa\|\mathbf{x}\|^2}$ and $\mathbf{I}_n$ is the $n$-dimensional identity matrix \textit{a.k.a.} Euclidean metric tensor. $\kappa<0$ is the sectional curvature of the manifold.

\subsubsection*{Gyrovector Addition}
Existing works \cite{ganea2018hyperbolic,shimizu2020hyperbolic,mathieu2019continuous} adopt the gyrovector space framework \cite{ungar2005analytic,ungar2008gyrovector} as a non-associative algebraic formalism for hyperbolic geometry. The gyrovector operation $\oplus_\kappa$is termed \textit{M\"obius addition}
\begin{align}
    &\mathbf{x} \oplus_\kappa \mathbf{y} := \frac{(1 - 2\kappa\innerproductcomma{\mathbf{x}}{\mathbf{y}} - \kappa\|\mathbf{y}\|^2)\mathbf{x} + (1 + \kappa\|\mathbf{x}\|^2)\mathbf{y}}{1 - 2\kappa\innerproductcomma{\mathbf{x}}{\mathbf{y}} + \kappa^2 \|\mathbf{x}\|^2 \|\mathbf{y}\|^2},
\end{align}
where $\mathbf{x},\mathbf{y}\in \mathbb{D}^n_\kappa$. The M\"obius addition defines the addition of two gyrovectors that preserves the summation on the manifold. The induced M\"obius subtraction $\ominus_\kappa$ is defined as $\mathbf{x} \ominus_\kappa \mathbf{y} = \mathbf{x} \oplus_\kappa (-\mathbf{y})$. 

\subsubsection*{Local Diffeomorphism}
A diffeomorphism is a map between manifolds which is differentiable and has a differentiable inverse. In a small enough neighborhood of any point $\mathbf{x}$ in hyperbolic space, we define \textit{exponential map} $\exp^\kappa_\mathbf{x}:\mathcal{T}_\mathbf{x}\mathbb{D}_\kappa \to \mathbb{D}_\kappa$ and its inverse \textit{logarithmic map} $\log_\mathbf{x}^\kappa:  \mathbb{D}_\kappa \to \mathcal{T}_\mathbf{x}\mathbb{D}_\kappa$ as local diffeomorphisms between hyperbolic manifold and tangent spaces
\begin{align}
    &\exp_\mathbf{x}^\kappa (\mathbf{v}) = \mathbf{x} \oplus_\kappa \left( \tanh (\sqrt{|\kappa|} \frac{\lambda_\mathbf{x}^\kappa \|\mathbf{v}\|}{2}) \frac{\mathbf{v}}{\sqrt{|\kappa|}\|\mathbf{v}\|}  \right),\label{eq:expmap}\\
    &\log_\mathbf{x}^\kappa (\mathbf{y}) = \frac{2 \tanh^{-1} (\sqrt{|\kappa|}\| -\mathbf{x} \oplus_\kappa \mathbf{y} \|) (-\mathbf{x} \oplus_\kappa \mathbf{y})}{\sqrt{|\kappa|}\lambda_\mathbf{x}^\kappa \| -\mathbf{x} \oplus_\kappa \mathbf{y} \|},
\end{align}

\subsubsection*{Gyrovector Multiplication}
Known as the \textit{M\"obius scalar multiplication}, $\otimes_\kappa$ specifies the product of a scalar $r$ and a gyrovector $\mathbf{x}\in \mathbb{D}^n_\kappa$. As specified in \cite{ganea2018hyperbolic}, the scalar multiplication can be acquired through diffeomorphisms
\begin{equation}
    r \otimes_\kappa \mathbf{x} = \exp_\mathbf{o}^\kappa (r \log_\mathbf{o}^\kappa (\mathbf{x})),
\label{eq:mscalar-multi-ext}
\end{equation}
Given matrix $\mathbf{W}\in \mathbb{R}^{m\times n}$, one can further extend Eq.~(\ref{eq:mscalar-multi-ext}) to \textit{matrix-vector multiplication}, formulated by
\begin{align}
    &\mathbf{W} \otimes_\kappa \mathbf{x} = \exp_\mathbf{o}^\kappa (\mathbf{W} \log_\mathbf{o}^\kappa (\mathbf{x})). \label{eq:mvector-multi}
\end{align}
Broadcasting \cite{paszke2019pytorch} further enables $\oplus_\kappa$ and $\otimes_\kappa$ on batched representations. 

\subsubsection*{Geodesic Length}
The \textit{geodesic} is the generalized straight line on a Riemannian manifold, the distance between two points is the \textit{geodesic length}. For $\mathbf{x}, \mathbf{y} \in \mathbb{D}_\kappa^n$, the analytical distance is formulated by
\begin{equation}
    d_\mathbb{D}^\kappa(\mathbf{x},\mathbf{y}) = \frac{2}{\sqrt{|\kappa|}}\tanh^{-1}\left( \sqrt{|\kappa|}\| - \mathbf{x} \oplus_\kappa \mathbf{y}  \| \right).
    \label{eq:poincare-distance}
\end{equation}

\subsubsection*{Parallel Transport}
Parallel Transport defines a way of transporting the local geometry along smooth curves that preserves the metric tensors. Specifically in hyperbolic space, the parallel transport $\mathcal{PT}_{\mathbf{x}\to \mathbf{y}}$ from point $\mathbf{x}\in \mathbb{D}^n_\kappa$ to $\mathbf{y}\in \mathbb{D}^n_\kappa$ is formulated as
\begin{align}
    &\mathcal{PT}_{\mathbf{x}\to\mathbf{y}}^\kappa (\mathbf{v}) = \frac{\lambda^\kappa_\mathbf{x}}{\lambda^\kappa_\mathbf{y}} \operatorname{gyr}[\mathbf{y}, \ominus_\kappa\mathbf{x}]\mathbf{v}, \quad \text{where}\\
    &\operatorname{gyr}[\mathbf{a}, \mathbf{b}]\mathbf{c} = \ominus_\kappa(\mathbf{a}\oplus_\kappa \mathbf{b})\oplus_\kappa(\mathbf{a}\oplus_\kappa(\mathbf{b}\oplus_\kappa\mathbf{c})),
\end{align}
which is essentially a map from  $\mathcal{T}_\mathbf{x}\mathcal{M}$ to $\mathcal{T}_\mathbf{y}\mathcal{M}$ that carries a vector $\mathbf{v}\in\mathcal{T}_\mathbf{x}\mathcal{M}$ along the geodesic
from $\mathbf{x}$ to $\mathbf{y}$.

%% file: Content/Appendices/AppC.tex
\section*{Appendix C: Implementation \& Algorithm}
\subsection*{Hyperbolic Encoder \& Decoder}
\subsubsection*{Encoder}
Given node feature $\mathbf{x}^E\in \mathbb{R}^{|\mathcal{V}|\times f}$, we first map the feature to hyperbolic space with curvature $\kappa_\mathrm{s}$, using exponential map
\begin{align}
    \mathbf{x}^H = \exp^{\kappa_\mathrm{s}}_{\mathbf{o}} (\mathbf{x}^E),
\end{align}
then use the hyperbolic neural network \cite{ganea2018hyperbolic} for feature transformation within the ball $\mathbb{D}_{\kappa_\mathrm{s}}^d$, specifically
\begin{align}
    &\text{weight multiplication: }\mathbf{z}' = \exp_{\mathbf{o}}^{\kappa_\mathrm{s}}( \log_{\mathbf{o}}^{\kappa_\mathrm{s}} (\mathbf{x}^H) \mathbf{W}),\label{eq:mobius-feature-transformation}\\
    &\text{bias translation: }\mathbf{z} = \exp_{\mathbf{z}'}^{\kappa_\mathrm{s}} (\mathcal{PT}_{\mathbf{o}\to \mathbf{z}'}^{\kappa_\mathrm{s}} (\log_\mathbf{o}^{\kappa_\mathrm{s}}(\mathbf{b}))),\label{eq:mobius-bias-addition},
\end{align}
where $\mathbf{W}\in \mathbb{R}^{f\times d}$ is the feature transformation matrix, and $\mathbf{b}\in \mathbb{D}_{\kappa_\mathrm{s}}^d$ is the manifold bias. Before feeding into PDE block, we use hyperbolic activation to re-adapt the curvature of manifold, given the activation function $\sigma$,
\begin{align}
    \mathbf{z}(0) = \sigma^{\kappa_\mathrm{s} \to \kappa_\mathrm{t}}(\mathbf{z}) = \exp_{\mathbf{o}}^{\kappa_\mathrm{t}}( \sigma(\log_{\mathbf{o}}^{\kappa_\mathrm{s}} (\mathbf{x}^H) )).
\end{align}
\subsubsection*{Decoder}
For link prediction and node classification, we employ the same decoders and losses as in \cite{chami2019hyperbolic}. Specifically, for link prediction, we compute the probability of edges using Fermi-Dirac decoder \cite{krioukov2010hyperbolic,nickel2017poincare}
\begin{equation}
    \mathbb{P}((i,j)\in \mathcal{E}\mid\mathbf{z}_i, \mathbf{z}_j) = \frac{1}{e^{\left(d_{\mathbb{D}}^2\left(\mathbf{z}_i, \mathbf{z}_j\right)-r\right) / t}+1},
\end{equation}
where $r$ and $t$ are hyperparameters. For node classification, we employ the same feature transformation as in Eq.~(\ref{eq:mobius-feature-transformation}-\ref{eq:mobius-bias-addition}) to transform the node embeddings to label dimension, then feed into cross entropy loss. 

\subsubsection*{Optimization}
For the Euclidean and hyperbolic parameters (\textit{e.g.} weights, curvatures) and (\textit{e.g.} biases), respectively, we employ Riemannian optimizers \cite{becigneul2018riemannian} standard adaptive optimizers \cite{kingma2014adam}.

\subsection*{Instantiating Local Diffusivity}
The local diffusivity can be implemented via graph attention \cite{velivckovic2017graph} or curve attention \cite{ye2019curvature}. In practice, we find the latter approach performs to have better performance within our framework. First we introduce some key concepts to obtain the Ollivier-Ricci Curvature (ORC)
\begin{definition}[Probability Measure]
    Given $\mathcal{G}=(\mathcal{V}, \mathcal{E})$, for a node $v\in \mathcal{V}$ and its neighbor $\mathcal{N}(v) = \{v_i, v_2, \dots, v_k\}$, for any probability mass $\alpha\in [0,1]$ of node $v$, the probability measure on $v$ is defined as
    \begin{equation}
        m_v^\alpha(v_i) = \begin{cases}
        \alpha, & \text{if } v_i =v\\
        \frac{1-\alpha}{k}, & \text{if } v_i \in \mathcal{N}(v)\\
        0, & \text{otherwise}
    \end{cases}
    \end{equation}
\end{definition}
\begin{definition}[Wasserstein Distance]
    Let $m^\alpha_u$, $m^\alpha_v$ be two probability measures around two end points $u, v\in \mathcal{V}$ of the edge $(u, v)\in \mathcal{E}$, the Wasserstein distance $W(m^\alpha_u, m^\alpha_v)$ seeks the optimal tranport plan between two two probability measures, which is given by solving the linear programming
\begin{align}
    &W(m^\alpha_u, m^\alpha_v) = \inf_{M} \sum_{u_i\in \mathcal{N}(u)}\sum_{v_j\in \mathcal{N}(v)} M(u_i, v_j) d(u_i, v_j),\nonumber\\
    & \begin{aligned}
        \text{s.t. }&\sum_{v_j\in \mathcal{N}(v)} M(u_i, v_j) = m_u^\alpha(u_i), \forall i, \\
    &\sum_{u_i\in \mathcal{N}(u)} M(u_i, v_j) = m_v^\alpha(v_j), \forall j,
    \end{aligned}
\end{align}
where $M(u_i, v_j):\mathcal{V}\times \mathcal{V} \to [0,1]$ is the transportation plan, \textit{i.e.,} the amount of probability mass transported from $u_i$ to $v_j$ along the shortest path $d(u_i, v_j)$.
\end{definition}
Then the coarse ORC \cite{ollivier2009ricci} $K(u,v)$ on edge $(u,v)$ is defined by comparing the Wasserstein distance $W(m_u, m_v)$ to the distance $d(u, v)$, formally,
\begin{align}
    K(u,v) = 1 - \frac{W(m_u, m_v)}{d(u, v)} \in (-2, 1).
\end{align}
With raw ORC values, we define the ORC-powered local diffusivity by
\begin{align}
    a^\mathrm{ldiff}_{i,j} = \mathrm{softmax}_{j\in \mathcal{N}(i)}\left(\operatorname{MLP}(K(i,j))\right),
\end{align}
where MLP is employed to make the curvature more adaptive to the manifold negative curvature $\kappa$. In our experiment, the $\operatorname{MLP}$ is instantiated as $\mathbb{R}\to \mathbb{R}^{d}\to \operatorname{LeakyReLU}\to \mathbb{R}^d$ with bias addition, where $d$ is the latent dim of embeddings throughout the diffusion process. We employ the channel-wise softmax that normalizes the outputs separately on each hidden dimension ranging from $1$ to $d$.

\begin{remark}
    A commonly posed question is: Is it applicable to employ the high complexity ORC approach in GNN which slows down the propagation? We answer in two folds: 1) The complexity of calculating the Ricci curvature exactly would require solving $|\mathcal{E}|$ LP problems but we could use approximation methods as advised in \cite{ye2019curvature} for parallel computation to speed up the computation. 2) the curvature values are only computed once as a preprocessing step, which is faster that LSPE encoding and is well-accepted by the literature \cite{ye2019curvature,topping2021understanding}.
\end{remark}

\subsection*{Instantiating Global Diffusivity}
In the setting when $\beta=1$, we fully disregard graph topology. Given the gradient matrix $\nabla\mathbf{z}(t)$ where each value element is computed by $(\nabla\mathbf{z}(t))_{i,j} = \log_{\mathbf{z}_i(t)}^{\kappa} (\mathbf{z}_j(t)) \in \mathbb{R}^d$. The global diffusivity $g_\phi$ seeks a permutation of directional diffusivity weights of all node pairs regardless of graph topology. Given the arbitrary time embedding $\mathbf{z}(t)$, we first project to the tangent space of north pole via $\log^\kappa_\mathbf{o}(\mathbf{z}(t))\in\mathbb{R}^{|\mathcal{V}|\times d}$, then the queries and keys can be computed by
\begin{align}
    \mathbf{q}(t) = \log^\kappa_\mathbf{o}(\mathbf{z}(t)) \mathbf{W}_\mathrm{q}, \quad \mathbf{k}(t) = \log^\kappa_\mathbf{o}(\mathbf{z}(t)) \mathbf{W}_\mathrm{k},
\end{align}
where $\mathbf{W}_\mathrm{q}, \mathbf{W}_\mathrm{k}\in\mathbb{R}^{d\times d}$ are trainable attention parameters. The unnormalized attention scores can be computed as
\begin{align}
    \mathbf{a}(t) = \sigma\left(\mathbf{q}(t) \left(\mathbf{k}(t)\right)^\top\right) \in \mathbb{R}^{|\mathcal{V}|\times |\mathcal{V}|},
\end{align}
where $\sigma$ is the sigmoid function. Then the re-normalized attention can is expressed as
\begin{align}
    \tilde{\mathbf{a}}(t) = (\mathbf{a}(t) \mathbf{1})^{-1} \mathbf{a}(t)  \in \mathbb{R}^{|\mathcal{V}|\times |\mathcal{V}|}.
\end{align}
Finally, the diffusivity $a^\mathrm{gdiff}_{i,j} = \tilde{\mathbf{a}}_{i,j}(t)$ is the $i,j$-th index of normalized weight $\tilde{\mathbf{a}}(t)$.
\subsubsection*{Multi-Head}
Let $H$ be the number of attention heads, a multi-head version can be achieved by changing $\mathbf{W}_\mathrm{q/k}\in \mathbb{R}^{d\times d}$ to $\mathbb{R}^{d\times (H\times d)}$, such that $\mathbf{q}(t)/\mathbf{k}(t)\in \mathbb{R}^{|\mathcal{V}|\times H\times d}$. The unnormalized attention scores can be computed as
\begin{align}
    \mathbf{a}_{i,j}(t) =\sigma \left(\mathbf{q}_{i,h}(t) \left(\mathbf{k}_{j,h}(t)\right)^\top\right)_{h=1}^{H} \in \mathbb{R}^{H}, 
\end{align}
To normalize the attention weights, first compute the sum of the attention weights for each node across all source nodes
\begin{align}
    \mathbf{r}_{i,h}(t) = \sum_{j=1}^{|\mathcal{V}|} \mathbf{a}_{i,j,h}(t) \in\mathbb{R}^{|\mathcal{V}|\times H}
\end{align}
Repeat the normalizer $|\mathcal{V}|$ times to match the shape of the attention weights
\begin{align}
    \mathbf{r}_{i,j,h}(t) =  \mathbf{r}_{i,h}(t), \quad\text{for $j=1,\dots,|\mathcal{V}|$}
\end{align}
such that the normalizer $\mathbf{r}(t)\in\mathbb{R}^{|\mathcal{V}|\times |\mathcal{V}|\times H}$. Normalize the attention weights by dividing them by the normalizer, and finall the head dim is averaged to get a scalar diffusivity
\begin{align}
    a^\mathrm{gdiff}_{i,j} = \tilde{\mathbf{a}}_{i,j}(t) = \frac{1}{H}\sum_{h=1}^H\frac{\mathbf{a}_{i,j,h}(t)}{\mathbf{r}_{i,j,h}(t)}.
\end{align}

\subsection*{HPDE Solver Algorithms}
The HEuler, HRK4 and HAM solver algorithms are respectively deligated to Alg.~\ref{alg:heuler}, Alg.~\ref{alg:hrk4} and Alg.~\ref{alg:ham}. 
\input{Content/Appendices/Algorithms}

\subsection*{Hyperparameter Settings}
For further implementation details, please refer to the attached code for settings.
\begin{table}[H]
\centering
\caption{Hyperparameter settings.}
\resizebox{0.9\linewidth}{!}{\input{table/parameter.tex}}
\label{tb:parameter-setting}
\end{table}

%% file: Content/Appendices/Algorithms.tex
\begin{algorithm}[H]
\caption{HEuler PDESolve}
\label{alg:heuler}
\textbf{Input}: Initial state $\mathbf{h}(0)\in \mathbb{D}_\kappa^d$, vector flow $\mathcal{F}_\theta^\kappa: \mathbb{D}_\kappa^d\to \mathbb{D}_\kappa^d$, time interval $[0,T]$, step size $\tau$.\\
\textbf{Parameter}: Diffusivity parameter $\theta$, sectional curvature $\kappa$. \\
\textbf{Output}: \text{Final state $\mathbf{h}(T)\in\mathbb{D}_\kappa^d$.}
\begin{algorithmic}[1] 
\STATE $t=0$
\FOR{$t<T$}
\STATE $\mathbf{k} = \mathcal{F}_\theta^\kappa(\mathbf{h}(t), t)$
\STATE $\mathcal{X}(t) = \log^\kappa_{\mathbf{h}(t)}(\mathbf{k})$
\STATE $\mathbf{h}(t+\tau) = \exp^\kappa_{\mathbf{h}(t)} (\tau \mathcal{X}(t))$
\IF{$t+\tau > T$}
\STATE $\delta = T-t$
\STATE $\mathbf{h}(T) = \exp_{\mathbf{h}(t)}^\kappa \left(\frac{\delta}{\tau} \log^\kappa_{\mathbf{h}(t)}(\mathbf{h}(t+\tau)) \right)$
\ENDIF
\STATE $t = t+\tau$
\ENDFOR
\STATE \textbf{return} $\mathbf{h}(T)$
\end{algorithmic}
\end{algorithm}

\begin{algorithm}[H]
\caption{HRK4 PDESolve}
\label{alg:hrk4}
\textbf{Input}: Initial state $\mathbf{h}(0)\in \mathbb{D}_\kappa^d$, vector flow $\mathcal{F}_\theta^\kappa: \mathbb{D}_\kappa^d\to \mathbb{D}_\kappa^d$, time interval $[0,T]$, step size $\tau$, coeffs $\{\phi\}=\{1, 3, 3, 1\}$.\\
\textbf{Parameter}: Diffusivity parameter $\theta$, sectional curvature $\kappa$. \\
\textbf{Output}: \text{Final state $\mathbf{h}(T)\in\mathbb{D}_\kappa^d$.}
\begin{algorithmic}[1] 
\STATE $\operatorname{normalize(\{\phi\})}$
\STATE $t=0$
\FOR{$t<T$}
\STATE $\mathcal{X}_{\mathbf{k}_1}(t) = \log^\kappa_{\mathbf{h}(t)}(\mathcal{F}_\theta^\kappa(\mathbf{h}(t), t))$
\STATE $\mathbf{k}_1 = \exp^\kappa_{\mathbf{h}(t)} (\tau \mathcal{X}_{\mathbf{k}_1}(t))$
\STATE $\mathcal{X}_{\mathbf{k}_2}(t) = \log^\kappa_{\mathbf{h}(t)}(\mathbf{k}_1)/3$
\STATE $\mathbf{k}_2 = \mathcal{F}_{\theta}^\kappa(\exp^\kappa_{\mathbf{h}(t)} (\tau \mathcal{X}_{\mathbf{k}_2}), t + \tau/3)$
\STATE $\mathcal{X}_{\mathbf{k}_3}(t) = \log^\kappa_{\mathbf{h}(t)}(\mathbf{k}_2) - \log^\kappa_{\mathbf{h}(t)}(\mathbf{k}_1)/3$
\STATE $\mathbf{k}_3 = \mathcal{F}_{\theta}^{\kappa}(\exp^\kappa_{\mathbf{h}(t)}(\tau \mathcal{X}_{\mathbf{k}_3}), t + 2\tau/3)$
\STATE $\mathcal{X}_{\mathbf{k}_4}(t) = \log^\kappa_{\mathbf{h}(t)}(\mathbf{k}_1) -  \log^\kappa_{\mathbf{h}(t)}(\mathbf{k}_2) + \log^\kappa_{\mathbf{h}(t)}(\mathbf{k}_3)$
\STATE $\mathbf{k}_4 = \mathcal{F}_{\theta}^\kappa (\exp^\kappa_{\mathbf{h}(t)}(\tau \mathcal{X}_{\mathbf{k}_4}), t+\tau)$
\STATE $\mathcal{X}_\mathrm{HRK4}(t) = \left(\sum_{i} \phi_i \log_{\mathbf{h}(t)}^\kappa (\mathbf{k}_i)\right)$
\STATE $\mathbf{h}(t+\tau) = \exp^\kappa_{\mathbf{h}(t)} (\tau \mathcal{X}_\mathrm{HRK4}(t))$
\IF{$t+\tau > T$}
\STATE $\delta = T-t$
\STATE $\mathbf{h}(T) = \exp_{\mathbf{h}(t)}^\kappa \left(\frac{\delta}{\tau} \log^\kappa_{\mathbf{h}(t)}(\mathbf{h}(t+\tau)) \right)$
\ENDIF
\STATE $t = t+\tau$
\ENDFOR
\STATE \textbf{return} $\mathbf{h}(T)$
\end{algorithmic}
\end{algorithm}

\begin{algorithm}[H]
\caption{HAM PDESolve}
\label{alg:ham}
\textbf{Input}: Initial state $\mathbf{h}(0)\in \mathbb{D}_\kappa^d$, vector flow $\mathcal{F}_\theta^\kappa: \mathbb{D}_\kappa^d\to \mathbb{D}_\kappa^d$, time interval $[0,T]$, step size $\tau$, min and max order $s_\mathrm{min}/s_\mathrm{max}$, Adam-Bashforth coeffs $\{\phi^\mathrm{HAB}\}$ and Adam-Moulton coeffs $\{\phi^\mathrm{HAM}\}$, double end queue $\mathbf{q}$.\\
\textbf{Parameter}: Diffusivity parameter $\theta$, sectional curvature $\kappa$. \\
\textbf{Output}: \text{Final state $\mathbf{h}(T)\in\mathbb{D}_\kappa^d$.}
\begin{algorithmic}[1] 
\STATE $\operatorname{normalize(\{\phi^\mathrm{HAB}\})}$
\STATE $\operatorname{normalize(\{\phi^\mathrm{HAM}\})}$
\STATE $\operatorname{init}(\mathbf{q})$
\STATE $i=0$
\FOR{$i\le s_\mathrm{min}$}
\STATE $\operatorname{pushhead}(\mathbf{q}, [\mathcal{X}_\mathrm{HRK4}(i\tau), \mathbf{h}_\mathrm{HRK4}(i\tau)])$
\STATE $i=i+1$
\ENDFOR
\STATE $t=\tau s_\mathrm{min}$
\FOR{$t<T$}
\STATE $\mathcal{X}_\mathrm{HAB}(t) = \sum_{j=1}^{\operatorname{len}(\mathbf{q})} \phi_j^\mathrm{HAB} \mathcal{PT}_{\mathbf{q}[i,1] \to \mathbf{h}(t)} \left(\mathbf{q}[j,0]\right)$
\STATE $\mathbf{h}^\star(t) = \exp_{\mathbf{h}(t)}^\kappa (\tau\mathcal{X}_\mathrm{HAB}(t))$
\STATE $\mathbf{h}_\mathrm{HAB}(t+\tau) = \mathcal{F}_\theta^\kappa (\mathbf{h}^\star(t), t+\tau)$
\STATE $\mathcal{X}_\mathrm{HAB}(t+\tau) = \log_{\mathbf{h}^\star(t)}^\kappa\left(\mathbf{h}_\mathrm{HAB}(t+\tau)\right)$
\STATE $\operatorname{pushhead}(\mathbf{q}, [\mathcal{X}_\mathrm{HAB}(t+\tau),\mathbf{h}_\mathrm{HAB}(t+\tau)])$
\STATE $\mathcal{X}_\mathrm{HAM}(t) = \sum_{j=1}^{\operatorname{len}(\mathbf{q})} \phi_j^\mathrm{HAM} \mathcal{PT}_{\mathbf{q}[i,1] \to \mathbf{h}(t)} \left(\mathbf{q}[j,0]\right)$
\STATE $\mathbf{h}(t+\tau) = \exp^\kappa_{\mathbf{h}(t)} (\tau \mathcal{X}_\mathrm{HAM}(t))$
\IF{$\operatorname{len}(\mathbf{q}) > s_\mathrm{max}$}
\STATE $\operatorname{poptail}(\mathbf{q})$
\ENDIF
\IF{$t+\tau > T$}
\STATE $\delta = T-t$
\STATE $\mathbf{h}(T) = \exp_{\mathbf{h}(t)}^\kappa \left(\frac{\delta}{\tau} \log^\kappa_{\mathbf{h}(t)}(\mathbf{h}(t+\tau)) \right)$
\ENDIF
\STATE $t = t + \tau$
\ENDFOR
\STATE \textbf{return} $\mathbf{h}(T)$
\end{algorithmic}
\end{algorithm}

%% file: table/parameter.tex
\begin{tabular}{@{}lllllllllll@{}}
\toprule
\textbf{Dataset}  & Time & $\tau$ & $\eta_1$ & $\eta_2$ & $\eta_3$ & int method & hidden dim & use lcc & weight decay & dropout \\ \midrule
\textbf{Cora}     & 8    & 1.0    & 1.0      & 0.6      & 0.1      & heuler     & 16         & False   & 1e-2         & 0.5     \\
\textbf{PubMed}   & 16   & 1.0    & 1.0      & 0.6      & 0.1      & hrk4       & 64         & True    & 5e-4         & 0.5     \\
\textbf{CiteSeer} & 16   & 1.0    & 1.0      & 0.8      & 0.8      & hrk4       & 64         & True    & 5e-4         & 0.5     \\
\textbf{Airport}  & 8    & 1.0    & 1.0      & 0.1      & 0.1      & heuler     & 64         & False   & 0.0          & 0.0     \\
\textbf{Disease}  & 4    & 1.0    & 1.0      & 0.4      & 0.1      & heuler     & 16         & False   & 5e-4         & 0.2     \\ \bottomrule
\end{tabular}

%% file: Content/Appendices/AppD.tex
\section*{Appendix D: Missing Proofs}

\subsection*{Proof of Proposition \ref{prop:proportional-geodesic-inter}}
\begin{proof}
Let $\mathbf{x}=\mathbf{h}(t)$, $\mathbf{y}=\mathbf{h}(t+\tau)$ and $\mathbf{z}=\mathbf{h}(t+\delta)$. We prove $\frac{d_\mathbb{D}^\kappa(\mathbf{h}(t), \mathbf{h}(t+\delta))}{ d_\mathbb{D}^\kappa(\mathbf{h}(t), \mathbf{h}(t+\tau)) }=\frac{\delta}{\tau}$ in the following. Starting from Eq.~(\ref{eq:hyperbolic-linear-interpolation}), according to Lemma 3 of \cite{ganea2018hyperbolic}, we have
\begin{align}
    \mathbf{z} & = \exp_\mathbf{x}^\kappa\left( \frac{\delta}{\tau} \log_\mathbf{x}^\kappa (\mathbf{y}) \right) = \gamma_{\mathbf{x}\to\mathbf{y}} \left(\frac{\delta}{\tau}\right) \\
    &= \mathbf{x} \oplus_\kappa \left( (-\mathbf{x} \oplus_\kappa \mathbf{y}) \otimes_\kappa \frac{\delta}{\tau}\right). \label{eq:proof-prop1-geodesic-exp}
\end{align}
Since $\frac{\delta}{\tau}\in (0,1)$, Eq.~(\ref{eq:proof-prop1-geodesic-exp}) meets the geodesic expression connecting $\mathbf{x}, \mathbf{y}\in \mathbb{D}^n_\kappa$ as shown in \cite{ungar2008gyrovector}. This proved $\mathbf{h}(t+\delta)$ is on the geodesic connecting $\mathbf{h}(t)$ and $\mathbf{h}(t+\tau)$. Let $\mathbf{u} = -\mathbf{x} \oplus_\kappa \mathbf{y} $, we further have
{\allowdisplaybreaks
\begin{align}
    &\frac{d_\mathbb{D}^\kappa(\mathbf{h}(t), \mathbf{h}(t+\delta))}{ d_\mathbb{D}^\kappa(\mathbf{h}(t), \mathbf{h}(t+\tau)) } =
    \frac{d_\mathbb{D}^\kappa (\mathbf{x}, \mathbf{z})}{d_\mathbb{D}^\kappa (\mathbf{x}, \mathbf{y})}\\
    &= \frac{ 
    \frac{2}{\sqrt{|\kappa|}} \tanh^{-1} (\sqrt{|\kappa|} \| -\mathbf{x}\oplus_\kappa \mathbf{z}\|)
    }{
    \frac{2}{\sqrt{|\kappa|}} \tanh^{-1}(\sqrt{|\kappa|} \| \mathbf{u} \| )
    }\\
    &= \frac{
    \tanh^{-1}( \sqrt{|\kappa|} \| -\mathbf{x}\oplus_\kappa \left( \mathbf{x} \oplus_\kappa \left( \mathbf{u} \otimes_\kappa \frac{\delta}{\tau}\right) \right)\| )
    }{
    \tanh^{-1}( \sqrt{|\kappa|} \| \mathbf{u} \| )
    }\\
    &= \frac{
    \tanh^{-1}( \sqrt{|\kappa|} \| \mathbf{u} \otimes_\kappa \frac{\delta}{\tau}\| )
    }{
    \tanh^{-1}( \sqrt{|\kappa|} \| \mathbf{u} \| )
    }\\
    &= \frac{
    \tanh^{-1}( \sqrt{|\kappa|} \| \frac{1}{\sqrt{|\kappa|}} \tanh( \frac{\delta}{\tau}  \tanh^{-1} (\sqrt{|\kappa|} \|\mathbf{u}\|) \frac{\mathbf{u}}{\|\mathbf{u}\|} )  \|  )
    }{
    \tanh^{-1}( \sqrt{|\kappa|} \| \mathbf{u} \| )
    }\\
    &= \frac{
    \frac{\delta}{\tau} \tanh^{-1} (\sqrt{|\kappa|} \|\mathbf{u}\|) \|\frac{\mathbf{u}}{\|\mathbf{u}\|}\|
    }{
    \tanh^{-1}( \sqrt{|\kappa|} \| \mathbf{u} \| )
    } = \frac{\delta}{\tau},
\end{align}
}
which concludes the proof.
\end{proof}